\documentclass{ecai} 
\usepackage{times}
\usepackage{soul}
\usepackage{url}
\usepackage[utf8]{inputenc}
\usepackage[small]{caption}
\usepackage{graphicx}
\usepackage{amsmath}
\usepackage{amssymb} % from ICLR version
\usepackage{amsthm}
\usepackage{booktabs}
\usepackage{algorithm}
\usepackage{algorithmic}
\usepackage[switch]{lineno}
\usepackage{paralist}
\usepackage{stfloats}
\usepackage{natbib}

\usepackage{amsfonts}

\def\eqref#1{equation~\ref{#1}}
\def\1{\bm{1}}

\DeclareMathAlphabet{\mathsfit}{\encodingdefault}{\sfdefault}{m}{sl}
\SetMathAlphabet{\mathsfit}{bold}{\encodingdefault}{\sfdefault}{bx}{n}

\DeclareMathOperator*{\argmax}{arg\,max}
\DeclareMathOperator*{\argmin}{arg\,min}

\usepackage{amsfonts}       % blackboard math symbols
\usepackage{nicefrac}       % compact symbols for 1/2, etc.
\usepackage{microtype}      % microtypography
\usepackage{optidef}
\usepackage{xcolor}
\usepackage{dsfont}
\usepackage{todonotes}
\usepackage{xcolor}
\usepackage{multirow}
\usepackage{bm} %,bbm}
\usepackage[most]{tcolorbox}
\DeclareMathOperator*{\minimize}{Minimize}

\usepackage{multicol}
\usepackage{url}

\newcommand{\gray}[1]{{\color{gray}{#1}}}

\newcommand{\cC}{\mathcal{C}} \newcommand{\cD}{\mathcal{D}}
 \newcommand{\cF}{\mathcal{F}}
 
 \newcommand{\cL}{\mathcal{L}}

 \newcommand{\cS}{\mathcal{S}}
 
\newcommand{\cZ}{\mathcal{Z}} 
 
 \newcommand{\cX}{\mathcal{X}}

\newcommand{\EE}{\mathbb{E}} \newcommand{\RR}{\mathbb{R}}

\usepackage{esvect}
\makeatletter
\newcommand{\oset}[3][0ex]{\mathrel{\mathop{#3}\limits^{\vbox to#1{\kern-2\ex@\hbox{$\scriptstyle#2$}\vss}}}}
\makeatother

\usepackage{letltxmacro}
\LetLtxMacro\orgvdots\vdots
\LetLtxMacro\orgddots\ddots

\definecolor{myblue}{RGB}{0, 51, 153}

\usepackage{newfloat}
\usepackage{listings}
\usepackage{wrapfig}
\usepackage{mdframed}
\usepackage{enumitem}
\usepackage{bibentry}

\newcommand{\pbox}[3]{%
  \begingroup
  \setlength{\fboxsep}{#1} % Set the padding
  \colorbox{#2}{$#3$}
  \endgroup
}

\newcommand{\BibTeX}{B\kern-.05em{\sc i\kern-.025em b}\kern-.08em\TeX}

\begin{document}
\begin{frontmatter}
\paperid{123} 

%%% Use this command to specify the title of your paper.

\title{%Predict-Then-Optimize by Proxy: 
Learning Joint Models of Prediction and Optimization}

%%% Use this combinations of commands to specify all authors of your 
%%% paper. Use \fnms{} and \snm{} to indicate everyone's first names 
%%% and surname. This will help the publisher with indexing the 
%%% proceedings. Please use a reasonable approximation in case your 
%%% name does not neatly split into "first names" and "surname".
%%% Specifying your ORCID digital identifier is optional. 
%%% Use the \thanks{} command to indicate one or more corresponding 
%%% authors and their email address(es). If so desired, you can specify
%%% author contributions using the \footnote{} command.

%\author[]{Anonyumous Authors}

%\author[]{
%James Kotary \\
%University of Virginia \\
%jk4pn@virginia.edu\And 
%Vincenzo Di Vito \\
%University of Virginia \\
%eda8pc@virginia.edu \And 
%Jacob Christopher \\
%University of Virginia \\
%csk4sr@virginia.edu \And
%Pascal Van Hentenryck \\
%Georgia Institute of Technology \\
%pvh@isye.gatech.edu \And 
%Ferdinando Fioretto \\
%University of Virginia \\
%fioretto@virginia.edu
%}

\author[A]{\fnms{James Kotary}\footnote{Corresponding authors with equal contribution. Emails: {jk4pn, eda8pc}@virginia.edu }}
\author[A]{\fnms{Vincenzo Di Vito}\footnotemark}
\author[A]{\fnms{Jacob Christopher}}
\author[B]{\fnms{Pascal Van Hentenryck}} 
\author[A]{\fnms{Ferdinando Fioretto}} 

\address[A]{University of Virginia}
\address[B]{Georgia Institute of Technology}

% \maketitle

\begin{abstract}
%Many real-world decision processes are modeled by optimization problems whose defining parameters are unknown and must be inferred from observable data. 
The Predict-Then-Optimize framework uses machine learning models to predict unknown parameters of an optimization problem from exogenous features before solving. This setting is common to many real-world decision processes, and recently it has been shown that decision quality can be substantially improved by solving and differentiating the optimization problem within an end-to-end training loop. %This approach enables new end-to-end training with loss functions defined directly on the resulting decisions. 
However, this approach requires significant computational effort in addition to handcrafted, problem-specific rules for backpropagation through the optimization step, challenging its applicability to a broad class of optimization problems. This paper proposes an alternative method, in which optimal solutions are learned directly from the observable features by joint predictive models. The approach is generic, and based on an adaptation of the Learning-to-Optimize paradigm, from which a rich variety of existing techniques can be employed. Experimental evaluations show the ability of several Learning-to-Optimize methods to provide efficient and accurate solutions to an array of challenging Predict-Then-Optimize problems.
\end{abstract}
\end{frontmatter}
\section{Introduction}
\label{sec:introduction}
The \emph{Predict-Then-Optimize} (PtO) framework models decision-making processes as optimization problems whose parameters are only partially known while the remaining, unknown, parameters must be estimated by a machine learning (ML) model. %The remaining parameters are unknown but correlated with a set of observable features from which a machine learning model must be trained to predict them. 
The predicted parameters complete the specification of an optimization problem which is then solved to produce a final decision. 
The problem is posed as estimating the solution $\bm{x}^\star(\zeta) \in \cX \subseteq \mathbb{R}^n$ of a \emph{parametric} optimization problem:
\begin{subequations}
    \label{eq:opt_generic}
    \begin{align}
        \label{eq:opt_generic_objective}
        \bm{x}^\star(\bm{\zeta}) = \argmin_{\bm{x}} &\;\; f(\bm{x}, \bm{\zeta})    \\
        \label{eq:opt_generic_constraints}
        \textsl{such that:} &\;\;    \bm{g}(\bm{x}) \leq 0, \;\;   \bm{h}(\bm{x}) = 0, 
    \end{align}
\end{subequations}
given that parameters $\bm \zeta \in \cC \subseteq \RR^p$ are unknown, but that a correlated set of observable values $\bm{z} \in \cZ$ are available. Here $f$ is an objective function, and $\bm g$ and $\bm h$ define the set of the problem's inequality and equality constraints. 
The combined prediction and optimization model is evaluated on the basis of the optimality of its downstream decisions, with respect to $f$ under its ground-truth problem parameters \citep{elmachtoub2020smart}. 
This setting is ubiquitous to many real-world applications confronting the task of decision-making under uncertainty, such as planning the shortest route in a city, determining optimal power generation schedules, or managing investment portfolios.
For example, a vehicle routing system may aim to minimize a rider's total commute time by solving a shortest-path optimization model (\ref{eq:opt_generic}) given knowledge of the transit times $\bm{\zeta}$ over each individual city block. In absence of that knowledge, it may be estimated by prediction models based on exogenous data $\bm{z}$, such as weather and traffic conditions. In this context, more accurately predicted transit times $\hat{\bm{\zeta}}$ tend to produce routing plans $\bm{x}^{\star}({\hat{\bm{\zeta}}})$ with shorter commutes, with respect to the true city-block transit times $\bm{\zeta}$.

\vspace{-2pt} 
However, direct training of predictions from observable features to problem parameters can generalize poorly with respect to the ground-truth optimality achieved by a subsequent decision model \citep{mandi2023decision,kotary2021end}. 
To address this challenge, \emph{End-to-end Predict-Then-Optimize} (EPO) \citep{elmachtoub2020smart} has emerged as a transformative paradigm in data-driven decision making, where predictive models are trained to directly minimize loss functions defined on the downstream optimal solutions $\bm{x}^{\star}({\hat{\bm{\zeta}}})$. 

On the other hand, EPO implementations present two key challenges: {\bf (i)} They require backpropagation through the solution of the optimization problem (\ref{eq:opt_generic}) as a function of its parameters for end-to-end training. The required backpropagation rules are highly dependent on the form of the optimization model and are typically derived by hand analytically for limited classes of models \citep{amos2019optnet,agrawal2019differentiable}. {\bf (ii)} Furthermore, difficult decision models involving nonconvex or discrete optimization may not admit well-defined backpropagation rules. 

% Despite its promise, the EPO paradigm currently faces two primary challenges: \emph{limited expressivity}, as existing methods predominantly concentrate on linear, convex, or mixed-integer-linear programs as decision programs, and \emph{speed}, since current techniques necessitate integrating a slow optimization solver within the training operations of an end-to-end model.

\begin{figure} %{r}{0.55\linewidth}
    \centering
    \includegraphics[width=\linewidth]{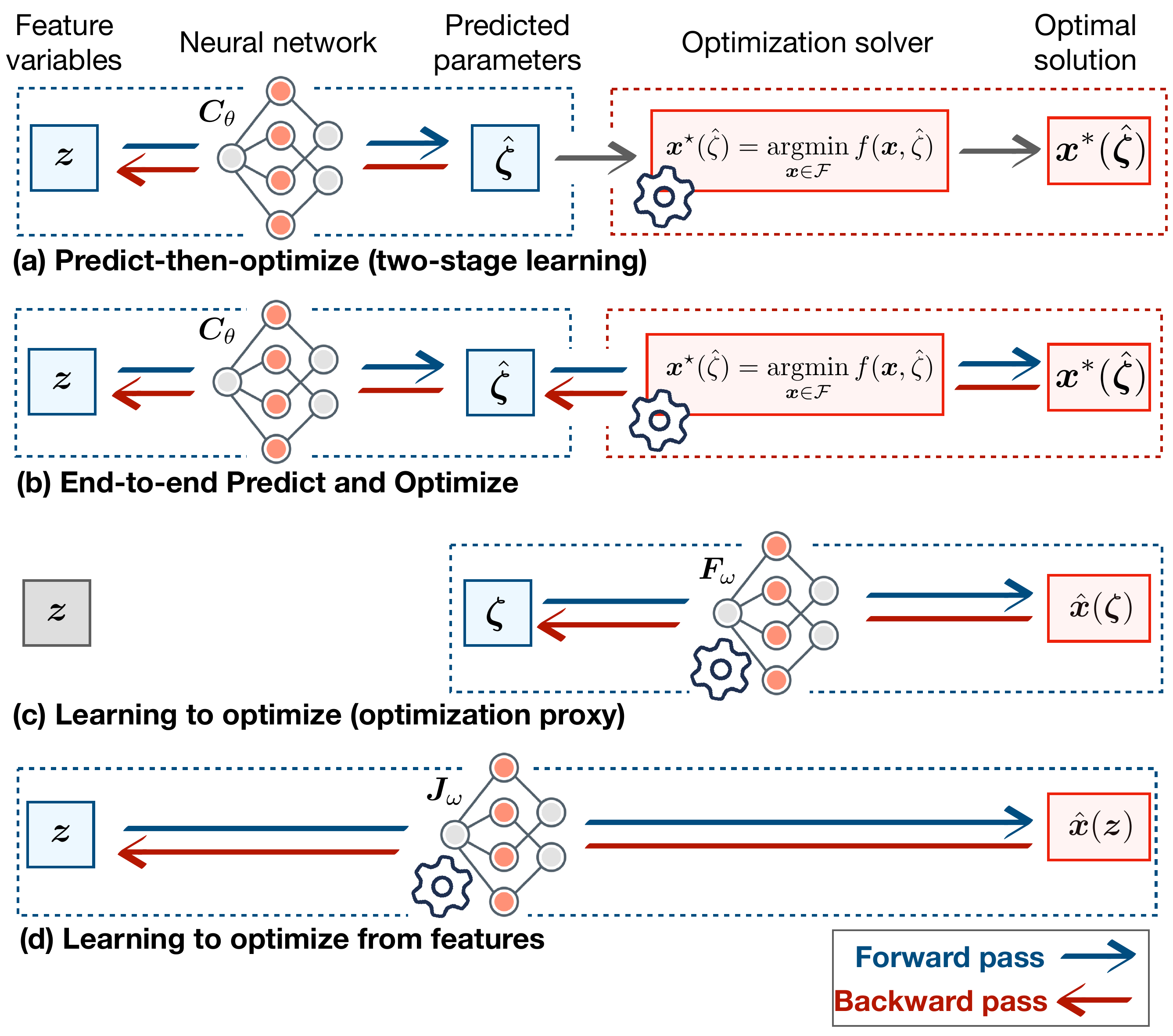}
    \vspace{-8pt}
    \caption{\small Illustration of Learning to Optimize from Features, in relation to other learning paradigms.}
    \vspace{15pt}
    \label{fig:LtOF_diagram}
\end{figure}

To address these challenges, this paper outlines a framework for training Predict-Then-Optimize models by techniques adapted from a separate but related area of work that combines constrained optimization end-to-end with machine learning. 
This paradigm, called \emph{Learn-to-Optimize} (LtO), learns a mapping between the parameters of an optimization problem and its corresponding optimal solutions using a deep neural network (DNN), as illustrated in Figure \ref{fig:LtOF_diagram}(c).

The resulting DNN mapping is then treated as an \emph{optimization proxy} whose role is to repeatedly solve difficult, but related optimization problems in real time \citep{vesselinova2020learning,fioretto2020lagrangian}. Several LtO methods specialize in training proxies to solve difficult problem forms, especially those involving nonconvex optimization. 

The methodology of this paper recognizes that existing LtO methods can provide an array of implementations for producing learned optimization proxies, which can handle hard optimization problem forms, have fast execution speeds, and are differentiable by construction. As such, they can be adapted to the Predict-Then-Optimize setting, offering an alternative to hard optimization solvers with handcrafted backpropagation rules. However, direct transfer of a pretrained optimization proxy into an EPO training loop leads to degradation of accuracy in the proxy solver, for which an end-to-end learning solution is proposed. The resulting \emph{Learning to Optimize from Features} (LtOF) framework extends the Learn-to-Optimize problem setting to \emph{encompass} that of Predict-Then-Optimize, by learning to construct optimal solutions directly from features, as illustrated in Figure \ref{fig:main-scheme}(d).
\iffalse
as shown in Section \ref{sec:EPOwithProxy}, due to the inability of LtO proxies to generalize outside their training distribution. To circumvent this distributional shift issue, this paper shows how to adapt the LtO methodology to learn optimal solutions directly from features.
\fi

\noindent \textbf{Contributions.} This paper makes the following novel contributions: \textbf{(1)} It investigates the use of pretrained LtO proxy models as a means to approximate the decision-making component of an EPO pipeline, and demonstrates a distributional shift effect between prediction and optimization models that leads to loss of accuracy in EPO training. \textbf{(2)} It proposes Learning to Optimize from Features (LtOF), in which existing LtO methods are adapted to learn solutions to optimization problems directly from observable features, circumventing the distributional shift effect over the problem parameters. \textbf{(3)} The generic LtOF framework is evaluated by adapting several well-known LtO methods to solve Predict-then-Optimize problems with difficult optimization components. Besides outperforming conventional two-stage approaches, \emph{the results show that difficult nonconvex optimization components can be incorporated into PtO pipelines naturally}, and illustrate how high decision quality can be reached in such cases even when gradient-based EPO alternatives fail.

\iffalse
\paragraph{Overview} The remaining sections are structured as follows: Section \ref{sec:PtO} gives a brief summary of related works on end-to-end learning frameworks for Predict-Then-Optimize, whose problem setting is shared by this paper. Section \ref{sec:LtO} gives an overview of existing Learn-to-Optimize frameworks, with emphasis on the collection of LtO methods which are adapted to solve PtO problems later in the paper. Section \ref{sec:Proxy_PtO} investigates the use of optimization proxies trained by those LtO methods, as surrogate differentiable optimization solvers in end-to-end training of PtO models. Section \ref{sec:LtOF} improves upon this idea by introducing Learning to Optimize from Features, and Section \ref{sec:Experiments} presents an experimental evaluation of LtOF, demonstrating its advantage in handling some difficult optimization forms under complex feature mappings when compared common baseline methods.
\fi

\section{Problem Setting and Background}
\label{sec:PtO}

%%%% OLD TEXT 
In the Predict-then-Optimize (PtO) setting, three distributions of data are assumed. Observable features $\bm{z} \sim \cZ$ are correlated with the \emph{unknown} coefficients $\bm{\zeta} \sim \cC$ of a parametric optimization problem (\ref{eq:opt_generic}), which in turn defines corresponding optimal solutions $\bm{x}^\star(\bm{\zeta}) \sim \cX$. We aim to learn optimal solutions $\bm{x}^\star(\bm{\zeta})$ to problem (\ref{eq:opt_generic}) without knowing the objective coefficients $\bm{\zeta}$, but knowing instead the correlated features $\bm{z}$. More precisely, the goal is to learn a mapping $\hat{\bm{x}}_{\theta}: \cZ \to \cX$ from observable features $\bm{z}$ to feasible solutions of (\ref{eq:opt_generic_constraints}), while aiming to optimize their objective value (\ref{eq:opt_generic_objective}) under the ground truth coefficients $\bm{\zeta}$. Assuming a joint distribution $(\bm{z}, \bm{\zeta}) \sim \Omega$, this can be expressed as
\begin{equation}\label{eq:epo_goal}
    \minimize_\theta\; \EE_{(\bm{z}, \bm{\zeta}) \sim \Omega}\left[
        f\left( \hat{\bm{x}}_{\theta}(\bm{z}), \bm{\zeta} \right) 
        \right] .
\end{equation}

A deterministic mapping $\bm{x}^{\star}: \cC \to \cX$ from problem coefficients to optimal solutions is defined by optimization problem (\ref{eq:opt_generic}), and can be implemented using any solution method which solves (\ref{eq:opt_generic}). As such, approaches to solving (\ref{eq:epo_goal}) are commonly based on framing $\hat{\bm{x}}_{\theta}$ as a composite prediction-and-optimization model $\bm{x}^{\star} \circ \bm{C}_\theta: \cZ \to \cX$, in which $ \bm{C}_\theta: \cZ \to \cC$ is a neural network trained to estimate problem coefficients $\bm{\zeta}$ from the observable features $\bm{z}$. The prediction and optimization components $\bm{C}_\theta$ and $\bm{x}^{\star}$  are called, respectively, the \emph{first} and \emph{second} stage models.

Two main approaches are typically used to train the predictive component $\hat{\bm{\zeta}} = \bm{C}_\theta(\bm{z})$, in order to realize the training goal (\ref{eq:epo_goal}): 

\noindent \textbf{$\blacksquare$ Two-stage Method.} A conventional approach to training the parameter prediction model $\hat{\bm{\zeta}} = \bm{C}_\theta(\bm{z})$ is the \emph{two-stage} method. It trains to predict the problem coefficients by MSE regression from their ground-truth values; i.e. with loss function
$\ell(\hat{\bm{\zeta}}, \bm{\zeta}) = \| \hat{\bm{\zeta}} - \bm{\zeta} \|_2^2$, without accounting for the downstream optimization during training. This direct minimization of prediction errors is consistent with the goal (\ref{eq:epo_goal}) of optimizing the objective value
$f(\bm{x}^\star(\hat{\bm{\zeta}}), \bm{\zeta})$. However, by ignoring the effect of error propagation from predicted coefficients $\hat{\bm{\zeta}}$ to downstream optimal solutions $\bm{x}^{\star}(\hat{\bm{\zeta}})$ in the second stage, this naive approach is known to result in suboptimal objective values $f(\bm{x}^\star(\hat{\bm{\zeta}}), \bm{\zeta})$ \citep{elmachtoub2020smart}, yielding poor performance on the PtO goal (\ref{eq:epo_goal}).

\noindent \textbf{$\blacksquare$ End-to-End Predict-Then-Optimize.}
Improving on the two-stage method, the End-to-end Predict-Then-Optimize (EPO) approach trains $\bm{C}_\theta$ directly to optimize the objective  $f(\bm{x}^\star(\hat{\bm{\zeta}}), \bm{\zeta})$ as a loss function by gradient descent. This is enabled by finding or approximating the derivatives through $\bm{x}^\star(\hat{\bm{\zeta}})$. This corresponds to end-to-end training of the PtO goal (\ref{eq:epo_goal}), where $\hat{\bm{x}}_{\theta} = \bm{x}^{\star} \circ \bm{C}_\theta$:
\begin{equation}\label{eq:epo_erm}
    \minimize_\theta\; \EE_{(\bm{z}, \bm{\zeta}) \sim \Omega}\left[
        f\left( \bm{x}^\star(\bm{C}_{\theta}(\bm{z})), \bm{\zeta} \right) 
        \right].
\end{equation}
\iffalse
Such an integrated training of prediction and optimization is referred to as \emph{Smart Predict-Then-Optimize} \citep{elmachtoub2020smart}, \emph{Decision-Focused Learning} \citep{wilder2018melding}, or {End-to-End Predict-Then-Optimize} (EPO) \citep{tang2022pyepo}. This paper adopts the latter term throughout, for consistency. 
\fi

EPO training consistently outperforms two-stage methods with respect to the goal (\ref{eq:epo_goal}), especially when the mapping $\bm{z} \to \bm{\zeta}$ is complex, due to the aforemention error propagation effect. See Figure \ref{fig:LtOF_diagram} (a) and (b) for an illustrative comparison, where the constraint set is denoted with $\cF$.  An overview of related work on EPO is reported in Section \ref{app:related_work}.

\subsection*{Challenges in End-to-End Predict-Then-Optimize} 
Despite their advantages over the two-stage, EPO methods are known to face two key challenges:
\textbf{(1)} \textbf{Differentiability}: the need for handcrafted backpropagation rules through $\bm{x}^{\star}(\bm{\zeta})$, which are highly dependent on the form of problem (\ref{eq:opt_generic}), and rely on the assumption of derivatives $ \frac{\partial \bm{x}^{\star}}{\partial \bm{\zeta}}$ which may not exist, and require that the mapping (\ref{eq:opt_generic}) is unique, producing a well-defined function; 
\textbf{(2)} \textbf{Efficiency}: the need to solve the optimization (\ref{eq:opt_generic}) to produce $\bm{x}^{\star}(\bm{\zeta})$ for each sample, in deployment and at each iteration of training. 
The results of Section \ref{sec:Experiments} demonstrate a \emph{further} potential pitfall of EPO training: even when differentiable, nonconvexity of (\ref{eq:opt_generic}) can cause its gradients to provide unhelpful descent directions for EPO training. 

This paper is motivated by a need to address these disadvantages. To do so, it recognizes a body of work on training DNNs as \emph{learned optimization proxies} which have fast execution, are automatically differentiable by design, and specialize in learning mappings $\bm{\zeta} \rightarrow \bm{x}^{\star}(\bm{\zeta})$ of hard optimization problems. 
While the next section discusses why the direct application of learned proxies as differentiable optimization solvers in an EPO approach tends to fail, Section \ref{sec:LtOF} presents a successful adaptation of the approach, in which predictive models are trained to solve the PtO training goal (\ref{eq:epo_goal}) \emph{directly}.

\section{EPO with Optimization Proxies}
\label{sec:EPOwithProxy}
The Learning-to-Optimize problem setting encompasses a variety of distinct methodologies with the common goal of learning to solve optimization problems. This section characterizes the LtO setting, before proceeding to describe an adaptation of LtO methods to the  Predict-Then-Optimize setting.

\noindent\textbf{$\blacksquare$ Learning to Optimize.}
The idea of training DNN models to emulate optimization solvers is referred to as \emph{Learn-to-Optimize (LtO)}  \citep{kotary2021end}. 
Here the goal is to learn a mapping $\bf{F}_\omega: \cC  \to \cX$ 
from the parameters $\bm{\zeta}$ of an optimization problem (\ref{eq:opt_generic}) to its corresponding optimal solution
$\bm{x}^\star(\bm{\zeta})$(see Figure \ref{fig:LtOF_diagram} (c)).
The resulting \emph{proxy} optimization model has as its learnable component a DNN denoted $\hat{\bm{F}}_{\omega}$,
which may be augmented with further operations $\bm{\cS}$ such as constraint corrections or unrolled solver steps, so that $\bf{F}_\omega = \bm{\cS} \circ \hat{\bm{F}}_{\omega}$. While training such a lightweight model to emulate optimization solvers is in general difficult, it is made tractable by restricting the task over a \emph{limited distribution} of problem parameters $\bm{\zeta} \sim \cC$.  
A variety of LtO methods have been proposed, many specializing in learning to solve problems of a specific form. % Some are based on supervised learning, in which case precomputed solutions $\bm{x}^{\star}(\bm{\zeta})$ are required as target data in addition to parameters $\bm{\zeta}$ for each sample.
Some are based on supervised learning, where precomputed solutions $\bm{x}^{\star}(\bm{\zeta})$ are required as target data in addition to parameters $\bm{\zeta}$ for each sample. Others are \emph{self-supervised}, requiring only knowledge of the problem form (\ref{eq:opt_generic}) along with instances of the parameters $\bm{\zeta}$ for supervision in training. 
LtO methods employ special learning objectives to train the proxy model $\bm{F}_\omega$: 
\begin{equation}\label{eq:LtO_goal}
    \minimize_{\omega}\; \EE_{\bm{\zeta} \sim \cC}
    \left[ \ell^{\text{LtO}} 
        \Big( 
            \bm{F}_\omega(\bm{\zeta}), \bm{\zeta}
        \Big) 
    \right],
\end{equation}
where $\ell^{\text{LtO}}$ represents a loss %, or a combination of loss and some post-processing method, 
that is specific to the LtO method employed.
A primary challenge in LtO is ensuring the satisfaction of constraints $\bm{g}(\hat{\bm{x}})\leq 0$ and $\bm{h}(\hat{\bm x}) = 0$  by the solutions $\hat{\bm{x}}$ of the proxy model $\bm{F}_\omega$. This can be achieved, exactly or approximately, by a variety of methods, for example iteratively retraining Equation~(\ref{eq:LtO_goal}) while applying dual optimization steps to a Lagrangian loss function \citep{fioretto2020lagrangian,park2023self}, or designing $\bm{\cS}$ to restore feasibility \citep{donti2021dc3}, as reviewed in Section \ref{app:lto_methods}. In cases where small constraint violations remain in the solutions $\hat{ \bf{x} }$ at inference time, they can be removed by post-processing with efficient projection or correction methods as deemed suitable for the particular application \citep{kotary2021end}.

\subsection*{EPO with Pretrained Optimization Proxies}
\label{sec:Proxy_PtO}

Viewed from the Predict-then-Optimize lens, learned optimization proxies have two beneficial features by design: {\bf (1)} they enable very fast solving times compared to conventional solvers, and {\bf (2)} are differentiable by virtue of being trained end-to-end. Thus, a natural question is  whether it is possible to use a pre-trained optimization proxy to substitute the differentiable optimization component of an EPO pipeline. Such an approach modifies the EPO objective (\ref{eq:epo_goal}) as:
\begin{equation}\label{eq:LtO_pretrained}
    \!\!\!\!\minimize_\theta\;
    \EE_{(\bm z, \bm{\zeta}) \sim \Omega} 
    \biggl[
    f\Bigl(
        \overbrace{ \pbox{1pt}{gray!20}{\bm{F}_\omega}
         \bigl( \underbrace{\bm{C}_\theta(\bm{z})}_{\hat{\bm{\zeta}}}}^{\hat{\bm x}}\bigr),
        \bm{\zeta}
    \Bigr)
    \biggr],
\end{equation}

\!\!where the solver output $\bm{x}^\star(\hat{\bm{\zeta}})$ of problem (\ref{eq:epo_goal}) is replaced by the prediction $\hat{\bm{x}}$ obtained by LtO model $\bm{F}_\omega$ on input $\hat{\bm{\zeta}}$ (gray highlights a \!\pbox{0pt}{gray!20}{\text{pretrained}}\! model, with frozen weights $\omega$).

%%%%%%%%%%%%%%%
However, a critical challenge in LtO lies in the inherent limitation that ML models act as reliable optimization proxies \emph{only within the distribution of inputs they are trained on}. This challenges the implementation of the idea of using pretrained LtOs as components of an end-to-end Predict-Then-Optimize model  as the weights \( \theta \) update during training, leading to continuously evolving inputs \( \bm{C}_{\theta}(\bm{z}) \) to the pretrained optimizer \pbox{1pt}{gray!20}{\bm{F}_{\omega}}. Thus, to ensure robust performance, \pbox{1pt}{gray!20}{\bm{F}_{\omega}} must generalize well across virtually any input during training. However, due to the dynamic nature of \( \theta \), there is an inevitable \emph{distribution shift} in the inputs to \pbox{1pt}{gray!20}{\bm{F}_{\omega}}, destabilizing the EPO training.

\begin{figure}%[11]{r}{0.5\linewidth}
    \centering
    \includegraphics[width=1.0\linewidth]{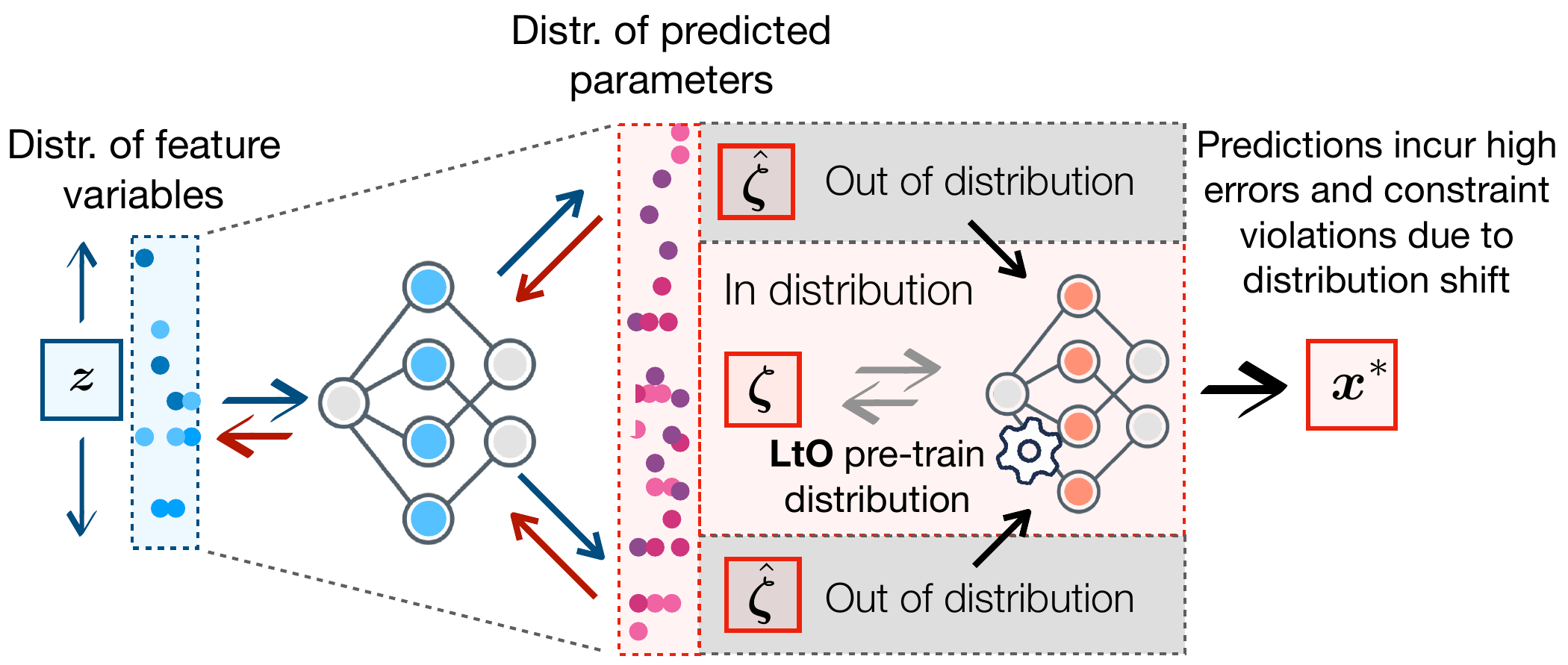}\\
    %\vspace{-10pt}
    \caption{\small Effect of shifting inputs received by the LtO proxy: a mismatch from its initial training distribution leads to inaccurate solutions when employed in PtO training.}
    \label{fig:main-scheme}
    \vspace{0.5cm}
\end{figure}

\begin{figure} %[11]{r}{0.48\linewidth}
    \centering
    \includegraphics[width=1.0\linewidth]{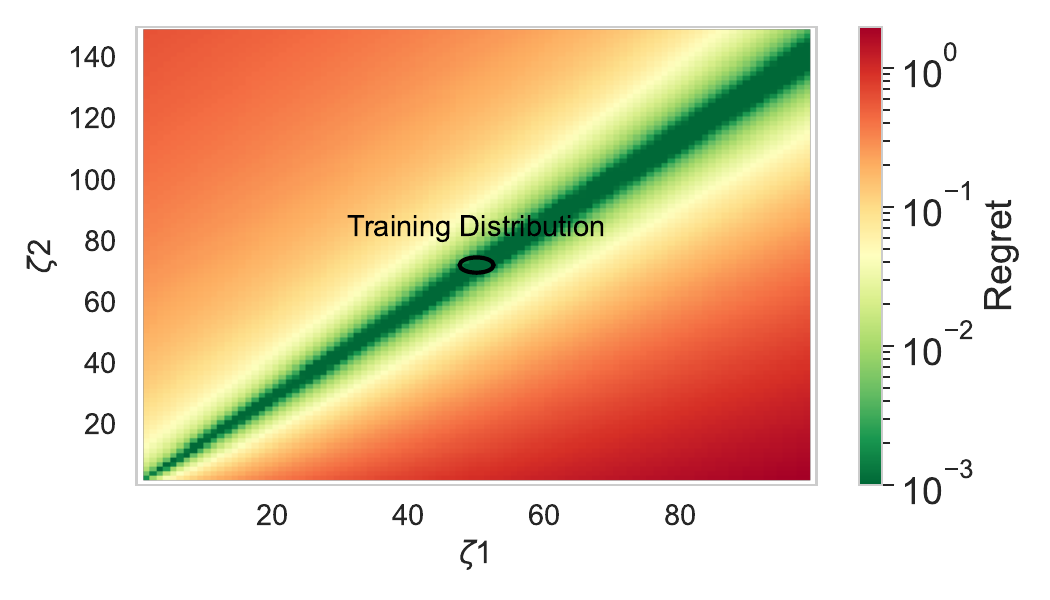}
    \vspace{-18pt}
    \caption{\small Effect on regret as LtO proxy acts outside its training set.}
    \vspace{10pt}
    \label{fig:distribution_shift}
    \vspace{0.5cm}
\end{figure}

Figures \ref{fig:main-scheme} and \ref{fig:distribution_shift} illustrate this issue. The former highlights how the input distribution to a pretrained proxy drifts during EPO training, impacting both output and backpropagation. The latter quantifies this behavior, exemplified on a simple two-dimensional problem (described in Appendix \ref{app:optimization_problems}), 
showing rapid increase in proxy regret as \( \hat{\bm{\zeta}} \) diverges from the initial training distribution \( \bm{\zeta} \sim \cC \) (in black). 
\footnote{The Appendix can be found online at \url{https://arxiv.org/pdf/2311.13087}.}

The experimental results of Table \ref{table:merged_regret_violation_table} reinforce these observations. While each proxy solver performs well within its training distribution, their effectiveness deteriorates sharply when utilized as in ~\eqref{eq:LtO_pretrained}. A step toward resolving this distribution shift issue allows the weights of ${\bm{F}_\omega}$ to adapt to its changing inputs, by \emph{jointly} training the prediction and optimization models:
\begin{equation}\label{eq:joint_dfl_goal}
    \minimize_{\theta, \omega} \;
    \EE_{(\bm z, \bm{\zeta}) \sim \Omega} 
    \biggl[
    f\Bigl(
        \overbrace{ {\bm{F}_\omega}
         \bigl( \underbrace{\bm{C}_\theta(\bm{z})}_{\hat{\bm{\zeta}}}}^{\hat{\bm x}}\bigr), 
        \bm{\zeta}
    \Bigr)
    \biggr].
\end{equation}
The predictive model $\bm{C}_\theta$ is then effectively absorbed into the predictive component of ${\bm{F}_\omega}$, resulting in a \emph{joint} prediction and optimization proxy model $\bm{J}_\phi = {\bm{F}_\omega} \circ \bm{C}_\theta$, where $\phi = (\omega, \theta)$.  Given the requirement for feasible solutions, the training objective (\ref{eq:joint_dfl_goal}) must be replaced with an LtO procedure that enforces the constraints on its outputs. This leads to the joint training framework presented next. 

\section{Learning to Optimize from Features}
\label{sec:LtOF}

The distribution shift effect described above arises due to the disconnect in training between the first-stage prediction network $\bm{C}_\theta : \cZ \rightarrow \cC$ and the second-stage optimization proxy $\bm{F}_\omega : \cC \rightarrow \cX$. 
However, the Predict-Then-Optimize setting (see Section \ref{sec:PtO}) ultimately only requires the combined model to produce a candidate optimal solution $\hat{\bm{x}} \in \cX$ given an observation of features $\bm{z} \in \cZ$. 
Thus, the intermediate prediction $\hat{\bm{\zeta}} = \bm{C}_\theta(\bm{z})$ in Equation~(\ref{eq:joint_dfl_goal}) is, in principle, not needed.  
This motivates the choice to learn direct mappings from features to optimal solutions of the second-stage decision problem. 
The joint model $\bm{J}_\phi : \cZ \to \cX $ %= \bm{C}_\theta \circ \bm{F}_\omega$} 
is trained by LtO procedures, employing
%The joint model $\bm{J}_\phi : \cZ \to \cX $ %= \bm{C}_\theta \circ \bm{F}_\omega$} is trained by Learning-to-Optimize procedures, employing
\begin{equation}
    \label{eq:LtOF_goal}
    \minimize_{\phi}\;
        \EE_{(\bm{z}, \bm{\zeta}) \sim \Omega} \biggl[
    \ell^{\text{LtO}}\Bigl( %(\bm{c}_\theta \circ \bar{\bm{F}}_\omega)
    \bm{J}_\phi(\bm{z}), \bm{\zeta} \Bigr)
    \biggr].
\end{equation}
This method can be seen as a generalization of the Learn-to-Optimize framework, to the Predict-then-Optimize setting. The key difference from the typical LtO setting is that problem parameters $\bm{\zeta} \in \cC$ are not known as inputs to the model, but the correlated features $\bm{z} \in \cZ$ are known instead. Therefore, estimated optimal solutions now take the form $\hat{\bm{x}} = \bm{J}_\phi(\bm{z})$ rather than $\hat{\bm{x}} =  \bm{F}_\omega(\bm{\zeta})$. Notably, this causes the self-supervised LtO methods to become \emph{supervised}, since the ground-truth parameters $\bm{\zeta} \in \cC$ now act only as target data while the feature variable  $\bm{z}$ takes the role of input data.

We refer to this approach as \emph{Learning to Optimize from Features (LtOF)}. Figure \ref{fig:LtOF_diagram} illustrates the key distinctions of LtOF relative to the other learning paradigms studied in this work. Figures (\ref{fig:LtOF_diagram}c) and (\ref{fig:LtOF_diagram}d) distinguish LtO from LtoF by a change in model's input space, from $\bm{\zeta} \in \cC$ to $\bm{z} \in \cZ$. This brings the framework into the same problem setting as that of the two-stage and end-to-end PtO approaches, shown in Figures (\ref{fig:LtOF_diagram}a) and (\ref{fig:LtOF_diagram}b). The key difference from the PtO approaches is that they produce an estimated optimal solution $\bm{x}^{\star}(\hat{\bm{\zeta}})$  by using a true optimization solver, but applied to an imperfect parametric prediction $\hat{\bm{\zeta}} = \bm{C}_\theta(\bm z)$. 
In contrast, LtOF directly estimates optimal solution $\hat{ \bm{x} }(\bm{z}) = \bm{J}_\phi(\bm{z})$ from features $\bm{z}$, circumventing the need to represent an estimate of $\bm{\zeta}$. 

\iffalse
\paragraph{Sources of Error in LtOF}
Like two-stage and EPO methods,  LtOF yields solutions subject to \( \emph{regret} \), which measures suboptimality relative to the true parameters \( \bm{\zeta} \), as defined in Equation \ref{eq:regret_def}. However, while in end-to-end and, especially, in the two-stage PtO approaches, the regret in $\bm{x}^{\star}(\hat{\bm{\zeta}})$ arises from imprecise parameter predictions $\hat{\bm{\zeta}} = \bm{C}_{\theta}(\bm{z})$ \citep{mandi2023decision}, 
in LtOF, the regret in the inferred solutions $\hat{\bm{x}}(\bm{z}) = \bm{J}_\phi(\bm{z})$ arises due to imperfect learning of the proxy optimization. This error is inherent to the LtO methodology used to train the joint prediction and optimization model $\bm{J}_{\phi}$, and persists even in typical LtO, in which $\bm{\zeta}$ are precisely known. In principle, a secondary source of error can arise from imperfect learning of the feature-to-parameter mapping $\bm{z} \to \bm{\zeta}$ within the joint model $\bm{J}_{\phi}$. However, these two sources of error are indistinguishable, as the prediction and optimization steps are learned jointly. \fi

\paragraph{Advantages of LtOF} Section \ref{sec:Experiments} demonstrates various LtOF implementations which can greatly outperform two-stage methods in terms of the optimality of their learned solutions, and are competitive with EPO training based on exact differentiation through $\bm{x}^{\star}(\bm{\zeta})$. In contrast to EPO, this is achieved \emph{without} access to exact optimization solvers, nor models of their derivatives. Two advantages of LtOF over EPO training, demonstrated in Section \ref{sec:Experiments}, are emphasized next. 
\subsection{Efficiency Benefits} 
Because the primary goal of the Learn-to-Optimize methodology is to achieve \emph{faster solving times} than are possible with conventional optimization solvers, the LtOF approach broadly inherits this advantage. As learned neural network mappings, LtOF models of joint prediction and optimization can have order-of-magnitude lower runtimes than other PtO methods which require to solve the full optimization problem (\ref{eq:opt_generic}) at inference time (see Table \ref{table:inference_times}). This enables the design of \emph{real-time} PtO models within the LtOF framework.

\subsection{Modeling Benefits} While EPO approaches require the use of problem-specific backpropagation rules, the LtOF framework instead requires an existing LtO implementation which can learn to solve the PtO problem's second-stage optimization component. Section \ref{sec:nonconvex_variant} shows how several LtOF implementations can succeed where EPO training fails, on a problem with a nonconvex oscillating objective term (see Figure \ref{fig:nc_results}). This result is significant but intuitive, since derivatives through a nonconvex function often do not correspond to useful descent directions for minimization. By contrast, the LtOF approach learns to construct solutions to the nonconvex problem directly from features, without relying on their derivatives, by drawing from existing LtO methods \citep{fioretto2020lagrangian,park2023self,donti2021dc3} that reliably learn to solve both convex and nonconvex optimization.

%The experiments of this paper focus on the scope of continuous optimization problems, whose LtO approaches share a common set of solution strategies. Notably, the next section illustrates cases in which EPO training fails due to nonconvexity of (\ref{eq:opt_generic}), while LtOF still results in high-quality solutions.

\section{Experimental Results}
\raggedbottom
\label{sec:Experiments}
The LtOF approach is evaluated against \emph{two-stage} and \emph{EPO} baselines, on three Predict-Then-Optimize tasks, each with a distinct second stage optimization component $\bm{x}^{\star}: \cC \rightarrow \cX$, as in \eqref{eq:opt_generic}. 
These include a convex quadratic program (QP), a nonconvex QP variant, and a nonconvex program with sinusoidal constraints, to showcase the flexibility of LtOF over various problem forms. 

\noindent\textbf{Performance Criteria.}
Each PtO method considered in this section is evaluated on the basis of its downstream decisions $\hat{\bm{x}}$, which are required to be \emph{feasible} to the problem constraints (\ref{eq:opt_generic_constraints}). Subject to feasibility, the object is to minimize the expected ground-truth objective $f(\hat{\bm{x}}, \bm{\zeta})$ as per (\ref{eq:epo_goal}). This is equivalent to minimizing expected \emph{regret}, defined as the magnitude of suboptimality of a solution $\hat{\bm{x}}$ to problem (\ref{eq:opt_generic}) with respect to the ground-truth parameters: 
\begin{equation}\label{eq:regret_def}
\textit{regret}(\hat{\bm{x}}, \bm{\zeta}) = f(\hat{\bm{x}}, \bm{\zeta}) - f(\bm{x}^\star(\bm{\zeta}), \bm{\zeta}).
\end{equation}
\noindent\textbf{LtOF methods.}
Three different LtOF implementations are evaluated on each PtO task, based on distinct Learn-to-Optimize methods, reviewed in detail in Section \ref{app:lto_methods}: 
\begin{itemize}[leftmargin=*, parsep=0pt, itemsep=0pt, topsep=0pt]
\item \emph{Lagrangian Dual Learning ({\bf LD})} \citep{fioretto2020lagrangian}, which augments a regression loss with penalty terms, updated to mimic a Lagrangian Dual ascent method to encourage the satisfaction of the problem's constraints.

\item \emph{Self-supervised Primal Dual Learning ({\bf PDL})} \citep{park2023self}, which uses an augmented Lagrangian function to perform joint self-supervised training of primal and dual networks for solution estimation.
% \item \emph{Self-supervised Primal Dual Learning ({\bf PDL})} \citep{park2023self}, which uses an augmented Lagrangian loss function to perform joint self-supervised training of primal and dual networks for solution estimation.
\item \emph{Deep Constraint Completion and Correction ({\bf DC3})} \citep{donti2021dc3}, which relies on a completion technique to enforce constraint satisfaction, while maximizing the empirical objective function in self-supervised training. 
\end{itemize}

While several other Learn-to-Optimize methods have been proposed in the literature, the above-described collection represents diverse subset which is used to demonstrate the potential of adapting the LtO methodology as a whole to the PtO setting. 

\iffalse
\paragraph{Metrics}.
In each task, observable feature data $\bm z \in \cZ$ are provided which correlate with the ground-truth optimization parameters $\bm{\zeta} \in \cC$. The goal is to learn a composite prediction and optimization model whose outputs $\hat{\bm{x}} \in \cX$ best optimize the empirical objective value $f(\hat{\bm{x}},\bm{\zeta})$.  
Where appropriate, we present the regret metric, which is equivalent but more informative. 

A strength of the LtOF approach is its generic handling of problem forms, which naturally accommodates unknown parameters in the constraints $g(\bm{x}, \bm{\zeta})\leq0$  and $h(\bm{x}, \bm{\zeta})=0$ as well as in the objective $f(\bm{x}, \bm{\zeta})$ (Equation \eqref{eq:opt_generic}). In the case where learned parameters $\bm{\zeta}$ affect the constraints of a problem, the magnitude of constraint violations under ground-truth $\bm{\zeta}$ is also compared across all methods under evaluation. 
\fi

\noindent \textbf{Feature generation}. End-to-End Predict-Then-Optimize methods integrate learning and optimization %to aim  %by using integrated optimization and learning methods, is 
to minimize the propagation of prediction errors--specifically, from feature mappings \( \bm{z} \to \bm{\zeta} \) to the resulting decisions \( \bm{x}^{\star}(\bm{\zeta}) \) (regret). It's crucial to recognize that \emph{even methods with high error propagation} can yield low regret \emph{if the prediction errors are low}. To account for this, EPO studies often employ synthetically generated feature mappings to control prediction task difficulty \citep{mandi2023decision}. 
Accordingly, for each experiment, we generate feature datasets \( (\bm{z}_1, \ldots \bm{z}_N) \in \mathcal{Z} \) from ground-truth parameter sets \( (\bm{\zeta}_1, \ldots \bm{\zeta}_N) \in \mathcal{C} \) using random mappings of increasing complexity. A feedforward neural network, \( \bm{G}^k \), initialized uniformly at random with \( k \) layers, serves as the feature generator \( \bm{z} = \bm{G}^k(\zeta) \). Evaluation is then carried out for each PtO task on feature datasets generated with \( k \in \{ 1,2,4,8\} \), keeping target parameters $\bm{\zeta}$ constant.

\noindent \textbf{Baselines}. In our experiments, LtOF models use feedforward networks with \( k \) hidden layers. For comparison, we also evaluate two-stage and, where applicable, EPO models, using architectures with \( m \) hidden layers for each  \( m \in \{ 1,2,4,8 \} \). Additionally, for each task and each \( m \in \{ 1,2,4,8 \} \), we evaluate EPO model with pretrained optimization proxies, which predictive model $\bm{C}_\theta$ is, or is not, pre-trained by MSE regression, as the \emph{two-stage} predictive model. Further training specifics are provided in Appendix \ref{app:experimental}.

\noindent \textbf{Comparison to LtO setting}. It is natural to ask how solution quality varies when transitioning from LtO to LtOF in a PtO setting, where solutions are learned directly from features. To address this question, each PtO experiment includes results from its analogous Learning to Optimize setting, where a DNN  $\bf{F}_\omega : \cC \to \cX$ learns a mapping from the parameters $\bm \zeta$ of an optimization problem to its corresponding solution $\bm{x}^\star(\bm{\zeta})$. This is denoted \( k\!=\!0 \; (\text{LtO}) \), indicating the absence of any feature mapping. 
All figures report the regret obtained by LtO methods for reference, although they are not directly comparable to the Predict-then-Optimize setting.

Finally, all reported results are averages across 20 random seeds and we refer the reader to Appendix \ref{app:experimental} for extensive additional details regarding experimental settings, architectures, data generation and hyperparamaters adopted.

\begin{figure}[t]
    \centering
    \begin{minipage}{0.48\textwidth}
        \centering
        \includegraphics[width=\linewidth]{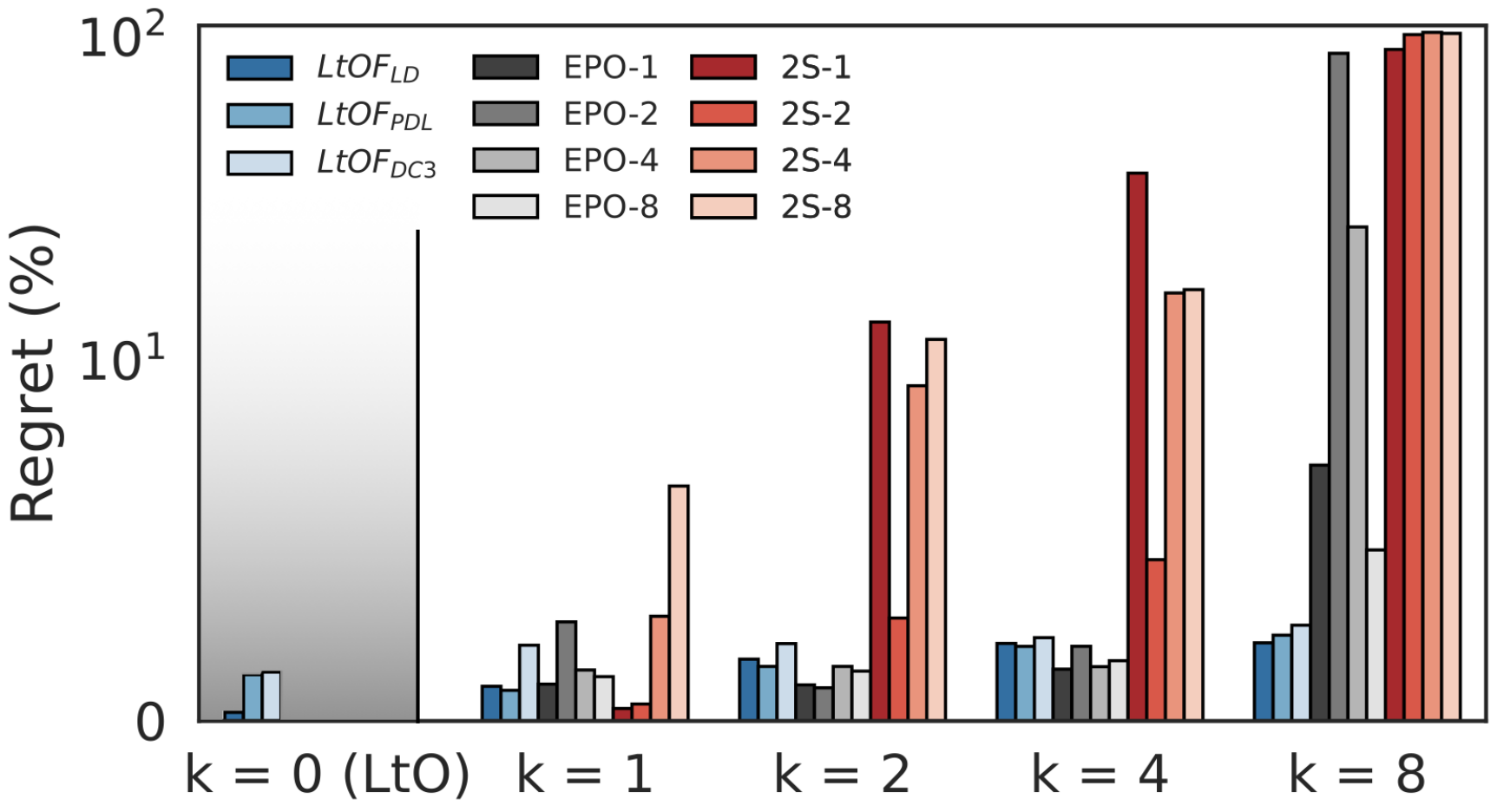}
        \vspace{-12pt}
        \caption{\small Comparison between \textbf{LtO} ($k\!=\!0$), \textbf{LtOF}, Two-stage (\textbf{2S}) and \textbf{EPO} ($k\!>\!1$) on the portfolio optimization. 2S(EPO)-$m$ indicates that the prediction model of the respective PtO method is an $m$ layer ReLU neural network. Plot y-axe is in semi log-scale.}
        \label{fig:portfolio_results}
    \end{minipage}
    \hfill
    \begin{minipage}{0.45\textwidth}
        \centering
        \vspace{10pt}
        \resizebox{\textwidth}{!}{%
        \begin{tabular}{clccc}
            \toprule
            & {\bf Method} & {\bf Portfolio} & {\bf N/conv.~QP} & {\bf AC-OPF} \\
            \midrule
            \multirow{4}{*}{\rotatebox{90}{\textbf{LtOF}}}
            % & LD it     & 0.0003014 & - & 0.00028 \\
            % & LD fct    & 0.0000002 & - & 0.15732 \\
            % & PDL it    & 0.0003200 & - & 0.00057 \\
            % & PDL fct   & 0.0000221 & - & 0.15130 \\
            % & DC3 it    & 0.0010897 & - & - \\
            % & DC3 fct   & 0.0002615 & - & - \\
            & LD it     & {\bf 0.0003} & 0.0000 & {\bf 0.0004} \\
            & \gray{LD fct}    & \gray{0.0000} & \gray{0.0032} 
            %\gray{0.1573$^{(1)}$} \\
            %&           &             &        
            & \gray{0.0516} \\ % Added row reporting Newton's method fct for OPF (LD)
            & PDL it    & {\bf 0.0003} & 0.0000 & 0.0006 \\
            & \gray{PDL fct}   & \gray{0.0000} & \gray{0.0032} %& \gray{0.1513$^{(1)}$} \\
            %&           &             &        
            & \gray{0.0207} \\ % Added row reporting Newton's method fct for OPF (PDL)
            & DC3 it    & 0.0011       & 0.0001 & - \\
            & \gray{DC3 fct}   & \gray{0.0003}  & \gray{0.0000} & - \\
            \midrule
            \multirow{4}{*}{\rotatebox{90}{\textbf{PtO}}}
            & PtO-1 et  & 0.0054 & 0.0122 & 2.5922 \\
            & PtO-2 et  & 0.0059 & 0.0104 & 2.5841 \\
            & PtO-4 et  & 0.0062 & 0.0123 & 2.5835 \\
            & PtO-8 et  & 0.0067 & 0.0133 & 2.5907 \\
            \bottomrule
        \end{tabular}
        }
        \vspace{0pt}
        \captionof{table}{\small Execution ({\it et}), inference ({\it it}), and feasibility correction ({\it fct}) times for {\bf LtOF} and {\bf PtO} (in seconds) for each sample. %For AC-OPF experiment, $^\star{(1)}$ denotes feasibility correction performed with Newton's method. 
        {\bf Two-stage} methods execution times are comparable to PtO's ones.}
        \vspace{15pt}
        \label{table:inference_times}
    \end{minipage}%
\end{figure}

%% RUNTIME IPOPT JUMP

%% PTO-1 et* 0.1729
%% PTO-2 et* 0.1645
%% PTO-3 et* 0.1777
%% PTO-4 et* 0.1651 

%%%%%%%%%%%%%%%%%%%%%%%%%%%%%%%%%%%%%%%%%%
\subsection{Convex Quadratic Optimization}
%%%%%%%%%%%%%%%%%%%%%%%%%%%%%%%%%%%%%%%%%%
\label{exp:CQO}
A well-known problem combining prediction and optimization is the Markowitz Portfolio Optimization \citep{rubinstein2002markowitz}. This task has as its optimization component a convex Quadratic Program (QP):
\begin{equation} %\tag{OPT}
    \label{eq:opt_portfolio}
        \bm{x}^{\star}(\bm{\zeta}) =  \argmax_{\bm{x} \geq \bm{0}} \; \bm{\zeta}^T \bm{x} - \lambda \bm{x}^T \bm{\Sigma} \bm{x}, \;\;\; \text{s.t. } \; \bm{1}^T \bm{x} = 1
\end{equation}
in which parameters $\bm{\zeta} \in \mathbb{R}^D$ represent future asset prices, and decisions $\bm{x} \in \mathbb{R}^D$ represent their fractional allocations within a portfolio. The objective is to maximize a balance of risk, as measured by the quadratic form covariance matrix $\Sigma$, and total return $\bm{\zeta}^T \bm{x}$. Historical prices of $D = 50$ assets are obtained from the Nasdaq online database \citep{NASDAQ} and used to form price vectors $\bm{\zeta}_i, \; 1\leq i \leq N$, with $N\!=\!12,000$ individual samples collected from 2015-2019. In the outputs $\hat{\bm{x}}$ of each LtOF method, any feasibility violations are restored, at \emph{low computational cost}, by first clipping $[\hat{\bm{x}}]_+$ to satisfy $\bm{x} \geq \bm{0}$, then dividing by its sum to satisfy $\bm{1}^T \bm{x} = 1$. The convex solver \texttt{cvxpy} \citep{diamond2016cvxpy} is used as the optimization component in each baseline method.

\noindent \textbf{Results}. Figure \ref{fig:portfolio_results} shows the percentage regret due to LtOF variants based on \textit{LD}, \textit{PDL} and \textit{DC3}. Two-stage and EPO models are evaluated for comparison,  with predictive components given various numbers of layers. For feature complexity $k>1$, each LtOF model outperforms the best two-stage model, increasingly with $k$ and up to nearly \emph{two orders of magnitude} when $k=8$.  
The EPO model, trained using exact derivatives  through  (\ref{eq:opt_portfolio}) using the differentiable solver in \texttt{cvxpylayers} \citep{agrawal2019differentiable} is competitive with LtOF until $k=4$, beyond which its best variant is outperformed by each LtOF variant. This shows the ability of LtOF models to reach high accuracy under complex feature mappings \emph{without} access to optimization solvers \emph{or} their derivatives, in training or inference, in contrast to EPO methods. 

Table \ref{table:inference_times} presents LtOF inference times (\( \textit{it} \)) and feasibility correction times (\( \textit{fct} \)), compared with the per-sample execution times (\( \textit{et} \)) for PtO methods. Run times for two-stage methods are  closely aligned with those of EPO, and thus omitted.
Notice how %the fastest 
LtOF methods are at least an order of magnitude faster than the PtO baselines. \iffalse This efficiency has two key implications: firstly, the per-sample speedup can significantly accelerate training for PtO problems. Secondly, it is especially advantageous during inference, particularly if data-driven decisions are needed in real-time.\fi

%%%%%%%%%%%%%%%%%%%%%%%%%%%%%%%%%%%%%%%%%%
\subsection{Nonconvex QP Variant}
\label{sec:nonconvex_variant}
%%%%%%%%%%%%%%%%%%%%%%%%%%%%%%%%%%%%%%%%%%

As a step in difficulty beyond convex QPs, this experiment considers a generic QP problem augmented with an oscillating objective term, resulting in a \emph{nonconvex} optimization problem:

\begin{subequations}
    \label{eq:bilinear}
    \begin{align}
        \label{eq:bilinear_obj}
        \mathbf{x}^{\star}(\bm{\zeta}) = \argmin_{\bm{x}} &\;\;
        \frac{1}{2} \bm{x}^T \bm{Q}  \bm{x} + \bm{\zeta}^T \sin( \bm{x} ) \\
        \texttt{s.t.} \;\; 
        & \bm{A} \bm{x} = \bm{b}, \; \bm{G} \bm{x} \leq \bm{h}.
    \end{align}
\end{subequations}
A variant of this formulation was used to evaluate the LtO methods proposed both in \cite{donti2021dc3} and in \cite{park2023self}. Following those works, $\bm{0} \preccurlyeq \bm{Q} \in \mathbb{R}^{n \times n}$, $\bm{A} \in \mathbb{R}^{n_{\text{eq}} \times n}$, $\bm{b} \in \mathbb{R}^{n_{\text{eq}}}$, $\bm{G} \in \mathbb{R}^{n_{\text{ineq}} \times n}$, $\bm{h} \in \mathbb{R}^{n_{\text{ineq}}}$ and each $\bm{\zeta_i}$ has elements drawn from a normal distribution.  The EPO baseline for comparison differentiates (\ref{eq:bilinear}) via the fixed-point conditions of a locally convergent Projected Gradient Descent method, implemented using the  \texttt{fold-opt} library \citep{kotary2023folded}; details on this EPO model are in Appendix \ref{app:experimental}. Feasibility is restored by a projection onto the feasible set, efficiently calculated by as a \emph{convex} QP. The problem dimensions are $n=50$ $n_{\text{eq}} = 25$, and $n_{\text{ineq}}=25$. 
% \iffalse
% QP problems whose objective Hessians are non-postive semidefinite, are known to be nonconvex and NP-Hard \cite{xia2020globally}. Such is the case for the  parametric \emph{bilinear programming} problem:
% \begin{subequations}
%     \label{eq:bilinear}
%     \begin{align*}
%         \mathbf{x}^{\star}(\mathbf{c}, \mathbf{d}) = \argmax_{\mathbf{0} \leq \mathbf{x}, \mathbf{y}  \leq \mathbf{1} } &\;\;
%         \mathbf{c}^T \mathbf{x} + \mathbf{d}^T \mathbf{y} + \lambda (\mathbf{x}^T \mathbf{Q} \mathbf{y} - \| \bm{x} \|^2   - \| \bm{y} \|^2 )  \\
%         \texttt{s.t.} \;\; 
%         & \sum \mathbf{x} = 1, \; \sum \mathbf{y} = 1,
%     \end{align*}
% \end{subequations}
% %  \left( \begin{matrix} \bm{x} \\ \bm{y}  \end{matrix} \right)
% which models optimal allocation of two resources $x$ and $y$, whose unknown linear objective functions are separable. The two allocation problems are couples only by the term $\mathbf{x}^T \mathbf{Q} \mathbf{y}$,  where $\bm{Q}$ consists uniform randomly sampled confounding factors, and the norm term amounts to a variance penalty which imparts smoothness to the mapping (\ref{eq:bilinear}). Unknown parameters $\bm{\zeta} = [\bm{c}, \bm{d}]$ are predicted from features, with ground-truth values sampled uniformly from $[0,1] \in \mathbb{R}^{20}$. 
% \fi

\begin{figure}[t]
        \centering
        \includegraphics[width=\linewidth]{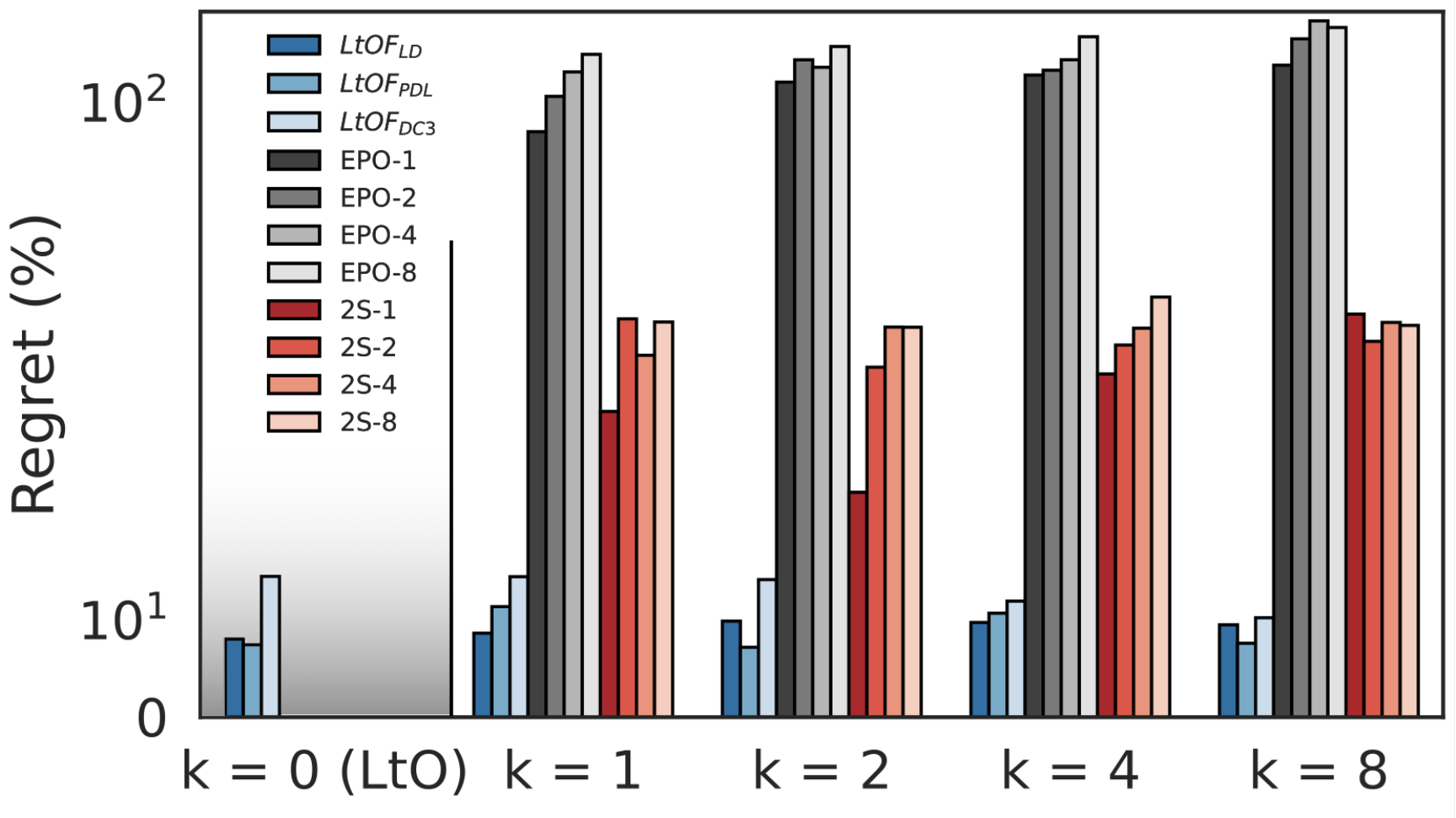}\\
        \hfill\includegraphics[width=0.95\linewidth]{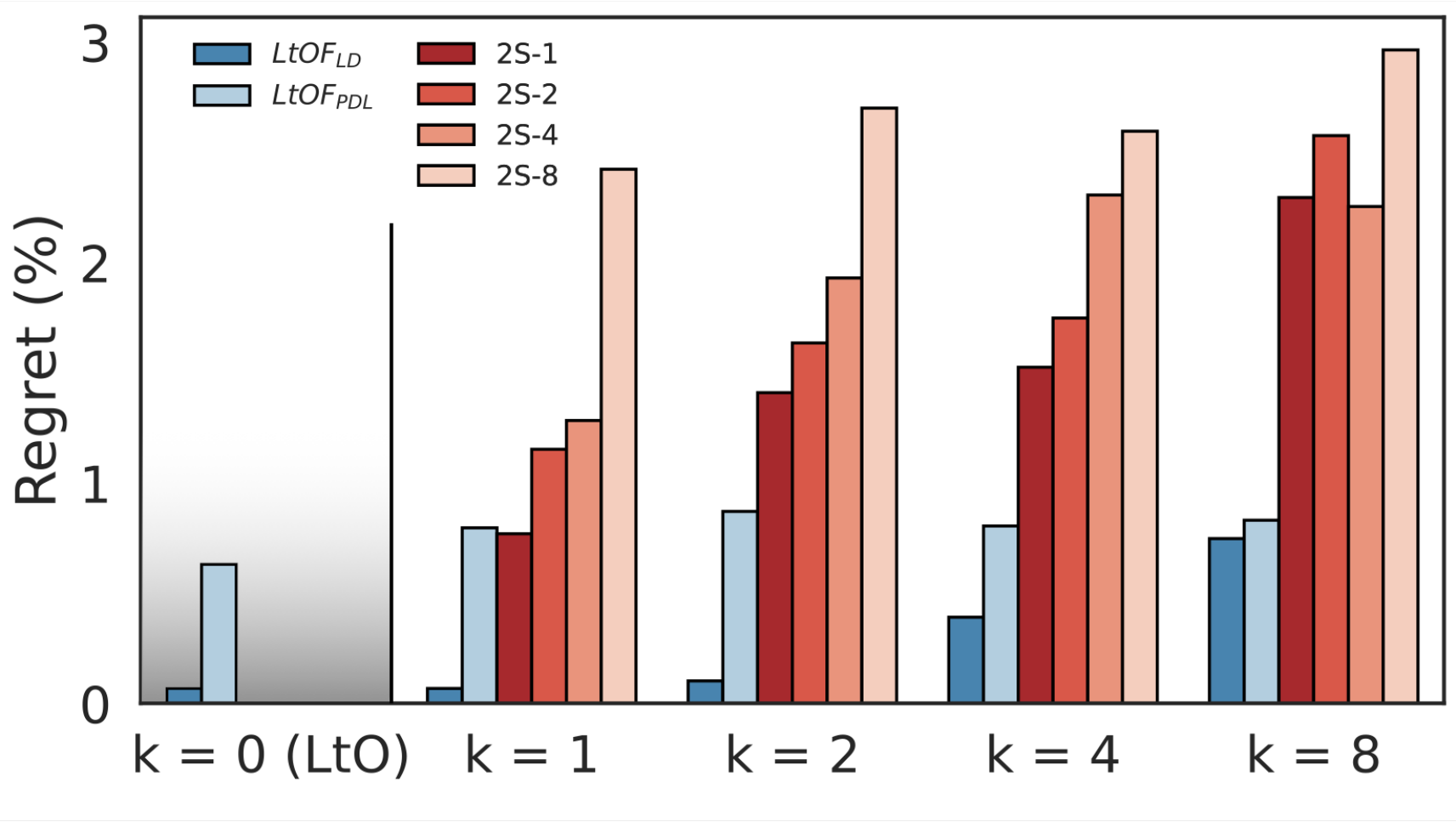}
        \vspace{-5pt}
        \caption{\small Comparison between LtO ($k=0$), LtOF, and Two Stage Method (2S) on the nonconvex QP (\textbf{top}) and 
        AC-OPF case (\textbf{bottom}). Top plot y-axe is in semi log-scale.}
        \vspace{15pt}
        \label{fig:nc_results}
\end{figure}

\noindent \textbf{Results.}
Figure \ref{fig:nc_results} (top) shows regret due to LtOF models based on  \textit{LD}, \textit{PDL} and \textit{DC3}, along with two baseline PtO methods. The best two-stage models perform poorly for all values of $k$, implying that the regret is particularly sensitive to prediction errors in the oscillating term. As alluded earlier, EPO training based on differentiation of (\ref{eq:bilinear}) performs even worse. This is intuitive, since gradients of the objective in (\ref{eq:bilinear_obj}) do not correspond to descent directions, due to its nonconvex $\sin$ term \citep{boydconvex}. Thus, backpropagation through (\ref{eq:bilinear}) guides the EPO model (\ref{eq:epo_erm}) to poor local minima in gradient descent training. The LtOF models achieve over $4\times$ times lower regret than the best baselines, suggesting strong potential for LtOF when predicting parameters of nonconvex objective functions. Additionally, the LtOF methods execute several times \emph{faster} than both baselines, after restoring feasibility.

\begin{table*}[t]
\centering
    \resizebox{\textwidth}{!}{%
    \begin{tabular}{clccc|ccc|ccc}
    \toprule
    & & \multicolumn{3}{c}{\bf Portfolio} & \multicolumn{3}{c}{\bf Nonconvex QP} & \multicolumn{3}{c}{\bf AC-OPF} \\
    \cmidrule(lr){3-5} \cmidrule(lr){6-8} \cmidrule(lr){9-11}
    & {\bf Method} & \(k=2\) & \(k=4\) & \(k=8\) & \(k=2\) & \(k=4\) & \(k=8\) & \(k=2\) & \(k=4\) & \(k=8\) \\
    \midrule
    \multirow{7}{*}{\rotatebox{90}{\bf LtOF}} &
      LD Regret       & 1.7170 & 2.1540 & {\bf 2.1700} & 9.9279 & {\bf 9.7879} & 9.5473  &{\bf 0.0748} & {\bf 0.3762} & {\bf 0.7231}\\
      %& {0.1016} & {0.4904} & {0.7470} \\
     %& LD Regret$^{(2)}$        &        &        &              &        &              &       & {\bf 0.0748} & {\bf 0.3762} & {\bf 0.7231} \\
    & \gray{LD Regret} (*)    & \gray{1.5739} & \gray{2.0903} & \gray{2.1386} & \gray{9.9250} & \gray{9.8211} & \gray{9.5556} & \gray{0.0013} & \gray{0.0071} & \gray{0.0195} \\
    & \gray{LD Violation} (*) & \gray{0.0010} & \gray{0.0091} & \gray{0.0044} & \gray{0.0148} & \gray{0.0162} & \gray{0.0195} & \gray{0.0020} & \gray{0.0037} & \gray{0.0042} \\
    & PDL Regret        & 1.5150 & 2.0720 & 2.3830 & {\bf 7.2699} & 10.747 & {\bf 7.6399} & 0.8714 & 0.8012 & 0.8373\\ 
    %& 0.9603 & 0.8543 & 0.8304 \\
    %& PDL Regret$^{(2)}$        &        &        &        &              &        &              & 0.8714 & 0.8012 & 0.8373 \\
    & \gray{PDL Regret} (*)   & \gray{1.4123} & \gray{1.9372} & \gray{2.0435} & \gray{7.2735} & \gray{10.749} & \gray{7.6394} & \gray{0.0260} & \gray{0.0243} & \gray{0.0242} \\
    & \gray{PDL Violation} (*)& \gray{0.0001} & \gray{0.0003} & \gray{0.0003} & \gray{0.0028} & \gray{0.0013} & \gray{0.0015} & \gray{0.0000} & \gray{0.0002} & \gray{0.0002} \\
    & DC3 Regret       & 2.1490 & 2.3140 & 2.6600 & 14.271 & 11.028 & 10.666 & - & - & - \\
    & \gray{DC3 Regret} (*)   & \gray{2.1490} & \gray{2.3140} & \gray{2.6600} & \gray{13.779} & \gray{11.755} & \gray{10.849} & \gray{-} & \gray{-} & \gray{-} \\
    & \gray{DC3 Violation} (*)& \gray{0.0000} & \gray{0.0000} & \gray{0.0000} & \gray{0.5158} & \gray{0.5113} & \gray{0.5192} & \gray{-} & \gray{-} & \gray{-} \\
    \midrule
    \multirow{2}{*}{\rotatebox{90}{\bf EPO}} 
    & EPO Regret (Best)         
                       & {\bf 0.9220} & {\bf 1.4393} & 4.7495        & 103.978      & 117.651  & 130.379    & -      & -      & -   \\
    & EPO w/ Proxy Regret (Best) 
                       & 154.40 & 119.31 & 114.69 & 812.75 & 804.26 & 789.50 & 389.04 & 413.89 & 404.74 \\
    & \multicolumn{1}{c}{\begin{tabular}[c]{@{}c@{}}EPO w/ Proxy and pretrained\\ prediction Regret (Best)\end{tabular}}  
                       & 14.783 & 23.052 & 22.158 & 214.53 & 198.02 & 241.93 & 52.344 & 55.195 & 60.052 \\

    \midrule
    & Two-Stage Regret (Best) 
                       & 2.8590 & 4.4790 & 91.326 & 36.168 & 37.399 & 38.297 & 1.4090 & 1.5280 & 2.4740 \\

    \bottomrule
    \end{tabular}
    }
    \caption{Percentage Regret and Constraint Violations for all experiments. (*) denotes ``Before Restoration''. Missing entries denote experiments not achievable with the associated method.}
    \label{table:merged_regret_violation_table}
\end{table*}

%%%%%%%%%%%%%%%%%%%%%%%%%%%%%%%%%%%%%%%%%%
\subsection{Nonconvex AC-Optimal Power Flow}
%%%%%%%%%%%%%%%%%%%%%%%%%%%%%%%%%%%%%%%%%%

Given a vector of marginal costs $\bm \zeta$ for each generator in an electrical grid, the AC-Optimal Power Flow problem optimizes the generation and dispatch of electrical power from generators to nodes with predefined demands. The objective is to minimize cost, while meeting demand exactly. The full optimization problem is specified in Appendix \ref{app:optimization_problems}, where a quadratic cost objective is minimized subject to nonconvex physical and engineering power systems constraints. This experiment simulates an energy market situation in which generation costs are as-yet unknown to the power system planners, and must be estimated based on correlated data. The goal is to predict costs so as to minimize cost-regret over an example network with $54$ generators, $99$ demand loads, and $118$ buses taken from the NESTA energy system test case archive \citep{coffrin2014nesta}. %Optimal solutions to the AC-OPF problem, are obtained using state-of-the-art Interior Point OPTimizer IPOPT \citep{ipopt}. 
Cost data are generated as perturbations around the test case's cost values as described in Appendix \ref{app:experimental}.
Feasibility is restored after LtOF by Newton's method on the constraint violations as specified in Appendix \ref{app:optimization_problems}.

\noindent \textbf{Results.} Figure \ref{fig:nc_results} (bottom) presents regret percentages, comparing LtOF to a two-stage baseline. Note that no general EPO exists for handling such nonconvex decision components. It is found empirically that the Lagrangian Hessian is not positive-semidefinite, preventing even a differentiable convex QP appromixation at its optima, as leveraged in \citep{wang2020scalable}. Further, the solving times reported in Table \ref{table:inference_times}, due to Casadi solver \cite{andersson2019casadi}, are prohibitively slow for EPO training.  {\it DC3} is also omitted, following \cite{park2023self}, due to its incompatibility with the LtO variant of this experiment. Notice how the \emph{best} two-stage model is outperformed by the {\it LD} variant of LtOF for $k>1$, and also by the {\it PDL} variant for $k>2$. Notably, PDL appears robust to increases in the feature mapping complexity $k$. On the other hand, it is outperformed in each case by the LD variant. {\em Notice how, in the complex feature mapping regime $k>1$, the best LtOF variant achieves up to an order of magnitude improvement in regret relative to the most competitive two-stage method.} 
Additionally, LtOF methods report orders-of-magnitude speed advantages in Table \ref{table:inference_times}.

Table \ref{table:merged_regret_violation_table} collects an abridged set of accuracy results due to each LtOF implementation and PtO baseline, across all experimental tasks. %Complete results can be found in Appendix \ref{app:experimental}.
In particular, average constraint violations and objective regret are shown in addition to regret after restoring feasibility. Infeasible results are shown in grey for purposes of comparing the regret loss due to restoration. 
Best results on each task are shown in bold. 
Additionally, regret achieved by the EPO framework with pretrained proxies (discussed in Section \ref{sec:Proxy_PtO}) are included. Average regrets between $10$ and $1000$ times higher than LtOF illustrate the effect of their distributional shifts on accuracy. 
Notice how, in the context of complex feature mappings $k \geq 2$, LtOF is competitive with  EPO, while bringing substantial computational advantages, and consistently outperforms two-stage methods, often, beyond an order of magnitude in regret.

\subsection{Learning to Optimize Methods}
\label{app:lto_methods}

Finally, this section describes  in more detail those LtO methods which were adapted to solve PtO problems by LtOF, in the above experiments.
Each description below assumes a DNN model $\cF$ and its weights $\omega$, which acts on problem parameters $\bm{\zeta}$ specifying an instance of problem (\ref{eq:opt_generic}), to produce an estimate of the optimal solution  $\hat{\bm{x}}\coloneqq \bm{F}_{\omega}(\bm{\zeta}) $, so that  $\hat{\bm{x}} \approx \bm{x}^{\star}(\bm{\zeta})$.

\paragraph{Lagrangian Dual Learning (LD).} 
\citet{fioretto2020lagrangian} constructs the following modified Lagrangian as a loss function for training the predictions  $\hat{\bm{x}} = \bm{F}_{\omega}(\bm{\zeta})$:
\begin{equation}
    \label{eq:ld_loss}
    \cL_{\textbf{LD}}(\hat{\bm{x}},\bm{\zeta}) =
        \| \hat{\bm{x}} - \bm{x}^{\star}(\bm{\zeta}) \|^2_2 +  \bm{\lambda}^T \left[ \bm{g}(\hat{\bm{x}},\bm{\zeta}) \right]_+ + \bm{\mu}^T  \bm{h}(\hat{\bm{x}},\bm{\zeta}).
\end{equation}

At each iteration of LD training, the model $\bm{F}_{\omega}$ is trained to minimize the loss $\cL_{\textbf{LD}}$. Then, updates to the multiplier vectors $\lambda$ and $\mu$ are calculated based on the average constraint violations incurred by the predictions $\hat{\bm{x}}$, mimicking a dual ascent method \cite{boyd2011distributed}. In this way, the method minimizes a balance of constraint violations and proximity to the precomputed target optima $\bm{x}^{\star}(\bm{\zeta})$.

\paragraph{Self-Supervised Primal-Dual Learning (PDL).} 
\citet{park2023self} use an augmented Lagrangian loss function 

\begin{equation}
\begin{split}
    \cL_{\textbf{PDL}}(\hat{\bm{x}},\bm{\zeta}) =
         f(\hat{\bm{x}}, \bm{\zeta}) + \hat{\bm{\lambda}}^T  \bm{g}(\hat{\bm{x}},\bm{\zeta})  + \hat{\bm{ \mu }}^T \bm{h}(\hat{\bm{x}},\bm{\zeta}) + \\ \frac{\rho}{2} \left( \sum_j \nu(g_j(\hat{\bm{x}})) + \sum_j \nu(h_j(\hat{\bm{x}}   )) \right),
\end{split}
\end{equation}
\label{eq:pdl_loss}

where $\nu$ measures the constraint violation. At each iteration of PDL training, a separate estimate of the Lagrange multipliers is stored for each problem instance, and updated by an augmented Lagrangian method \cite{boyd2011distributed} after training  $\hat{\bm{x}} = \bm{F}_{\omega}(\bm{\zeta}) $ to minimize \eqref{eq:pdl_loss}. In addition to the primal network $\bm{F}_{\omega}$, a dual network $\cD_{\bm{\zeta}}$ learns to store updates of the multipliers for each instance, and predict them as $(\hat{\bm{\lambda}}, \hat{\bm{\mu}}) = \cD_{\bm{\zeta}}(\bm{\zeta})$ to the next iteration. The method is self-supervised, requiring no precomputation of target solutions for training.

\paragraph{Deep Constraint Completion and Correction (DC3).} 
\citet{donti2021dc3} use the loss function
\begin{equation}
    \label{eq:dc3_loss}
    \cL_{\textbf{DC3}}(\hat{\bm{x}},\bm{\zeta}) =
         f(\hat{\bm{x}}, \bm{\zeta}) + \lambda \| \left[ \bm{g}(\hat{\bm{x}},\bm{\zeta}) \right]_+ \|_2^2 +  \mu  \| \bm{h}(\hat{\bm{x}},\bm{\zeta}) \|_2^2
\end{equation}
which combines a problem's objective value with two additional terms which aggregate the total violations of its equality and inequality constraints. The scalar multipliers $\lambda$ and $\mu$ are not adjusted during training. However, feasibility of predicted solutions is enforced by treating $\hat{\bm{x}} =  \hat{\bm{F}}_{\omega}(\bm{\zeta})$ as an estimate for only a subset of optimization variables. The remaining variables are completed by solving the underdetermined equality constraints $\bm{h}(\bm{x})=\bm{0}$ as a system of equations. Inequality violations are corrected by gradient descent on the their aggregated values  $\| \left[ \bm{g}(\hat{\bm{x}},\bm{\zeta}) \right]_+ \|^2$ . These completion and correction steps form the function $\bm{S}$, where $\bm{F}_{\omega}(\bm{\zeta}) = \bm{S} \circ \hat{\bm{F}}_{\omega}(\bm{\zeta})$.

\section{Related Work}
\label{app:related_work}

This section gives an overview of related work in the  Predict-Then-Optimize  setting. While the idea is general and has broader applications, differentiation through the optimization of (\ref{eq:opt_generic}) is central to EPO approaches.   Backpropagation of parametric quadratic programming problems was introduced by \cite{amos2019optnet}, which  implicitly differentiates the solution via its KKT conditions of optimality \cite{boydconvex}. \citet{agrawal2019differentiating} followed by proposing a differentiable cone programming solver, which uses implicit differentiation of problem-specific optimality conditions. That framework is leveraged by \cite{agrawal2019differentiable} to solve and differentiate general convex programs, by pairing it with a symbolic system for conversion of convex programs to canonical convex cone programs.

For many practical problems with discrete structure, such as linear programs, the mapping defined by (\ref{eq:opt_generic}) does not have well-defined derivatives, necessitating a suitable approximation in EPO training. \citep{elmachtoub2020smart, mandi2020smart} propose a surrogate loss function for (\ref{eq:epo_goal}) in cases where $f$ is linear, which admits useful subgradients for stochastic gradient descent training. \citep{wilder2018melding}  proposes backpropagation through linear programs by adding a smooth quadratic term to the objective and differentiating the resulting QP problem via \cite{amos2019optnet}.  \citep{berthet2020learning} also propose backpropagation through linear programs but by smoothing the mapping (\ref{eq:opt_generic}) through  random noise perturbations to the objective function. \citep{vlastelica2020differentiation} form approximate derivatives through linear optimization of discrete variables, by using a finite difference approximation between a pair of solutions with perturbed input parameters. Finally, \citep{mandi2020interior} use the barrier function of an interior point method with early stopping to provide a smoothed surrogate model for differentiable linear programming. 

A few recent works have addressed the topic of PtO learning without computing or approximating derivatives through optimization. In \cite{shah2022decision}, \cite{shah2024leaving}, \cite{zharmagambetov2024landscape} neural networks are trained to function as surrogate models of solution regret, for use in EPO training. In \cite{mandi2022decision}, PtO is recast as a learning to rank (LTR) problem and solved with various LTR methods, in favor of EPO training. 

\section{Discussion and Conclusions}

While the typical role of Learning to Optimize is to accelerate the solution of optimization problems, this paper demonstrates a novel use case: solving problems in the Predict-Then-Optimize scope. The adaptations of LtO described in this paper bring distinct advantages in the PtO setting, including real-time inference and enhanced ability to handle some PtO problems with nonconvex optimization. 

Another advantage of the Learning to Optimize from Features approach to PtO settings is its generic framework, which enables it to leverage a variety of existing techniques from the LtO literature. On the other hand, as such, a particular implementation of LtOF may inherit any limitations of the specific LtO method that it adopts. Future work should focus on understanding to what extent a broader variety of LtO methods can be applied to PtO settings; given the large variety of existing works in the area, such a task is beyond the scope of this paper. 
In particular, this paper does not investigate of the use of \emph{combinatorial} optimization proxies in learning to optimize from features. Such methods tend to use a distinct set of approaches from those studied in this paper, often relying on training by reinforcement learning \citep{bello2017neural,kool2019attention,mao2019learning}, and are not suited for capturing broad classes of optimization problems. As such, this direction is left to future work. 

One \emph{disadvantage} inherent to LtOF, compared to EPO, is the inability to recover parameter estimations from the predictive model, since optimal solutions are predicted directly from features. Although it is not required for the PtO problem setting, this may pose a challenge if transferring the learned parameters to external solvers is desirable. Furthermore, LtOF cannot be applied to PtO problems whose optimization component does not have an effective LtO solution.

By showing that effective Predict-Then-Optimize models can consist solely of Learning-to-Optimize methods, this paper has aimed to provide a unifying perspective on these as-yet distinct problem settings. The flexibility of its approach has been demonstrated by showing superior performance over PtO baselines with diverse problem forms. As the advantages of LtO are often best realized in combination with application-specific techniques, it is hoped that future work can build on these findings to maximize the practical benefits offered by LtO in settings that require data-driven decision-making.

\section{Acknowledgements}

This research is partially supported by NSF grants RI-2007164, RI-2232054, EPCN-2242931, AI Institute-2112533, and CAREER-2143706. 
%Fioretto is also supported by an Amazon Research Award and a Google Research Scholar Award. 
Its views and conclusions are those of the authors only.

\bibliography{mybibfile}

\appendix

\section{Optimization Problems}
\label{app:optimization_problems}

\paragraph{Illustrative $2D$ example}
Used for illustration purposes, the $2D$ optimization problem used to produce the results of Figure \ref{fig:distribution_shift} takes the form
\begin{subequations}
    \label{eq:dist_shift_fig_dummy_prob}
    \begin{align*}
        \mathbf{x}^{\star}(\bm{\zeta}) = \argmin_{\bm{x} } &\;\;
        \zeta_1 x_1^2 + \zeta_2 x_2^2\\
        \text{s.t.} \;\; 
        & x_1 + 2x_2 \leq 0.5, \\
        & 2x_1 - x_2 \leq 0.2, \\
        & x_1 + x_2 \leq 0.3
    \end{align*}
\end{subequations}
and its optimization proxy model is learned using \emph{PDL} training.

\paragraph{AC-Optimal Power Flow Problem.}
The OPF determines the least-cost
generator dispatch that meets the load (demand) in a power network.
The OPF is defined in terms of complex numbers, i.e., \emph{powers} of the form $S \!=\! (p \!+\! jq)$, where $p$ and $q$ denote active and reactive powers and $j$ the imaginary unit, \emph{admittances} of the form $Y \!=\! (g \!+\! jb)$, where $g$ and $b$ denote the conductance and susceptance, and \emph{voltages} of the form $V \!=\! (v \angle \theta)$, with magnitude $v$ and phase angle $\theta$. A power network
is viewed as a graph $({\cal N}, {\cal E})$ where the nodes $\cal N$
represent the set of \emph{buses} and the edges $\cal E$ represent
the set of \emph{transmission lines}. The OPF constraints include
physical and engineering constraints, which are captured in the AC-OPF
formulation of Figure \ref{model:ac_opf}.  The model uses $p^g$, and
$p^d$ to denote, respectively, the vectors of active power generation
and load associated with each bus and $p^f$ to describe the vector of
active power flows associated with each transmission line. Similar
notations are used to denote the vectors of reactive power $q$.
Finally, the model uses $v$ and $\theta$ to describe the vectors of
voltage magnitude and angles associated with each bus. The OPF takes
as inputs the loads $(\bm{p}^d\!, \bm{q}^d)$ and the admittance matrix
$\bm{Y}$, with entries $\bm{g}_{ij}$ and $\bm{b}_{ij}$ for each line
$(ij) \!\in\!  {\cal E}$; It returns the active power vector $p^g$ of
the generators, as well the voltage magnitude $v$ at the generator
buses. The problem objective \eqref{c_2a} captures the cost of the
generator dispatch and is typically expressed as a quadratic
function. Constraints \eqref{c_2b} and \eqref{c_2c} restrict the
voltage magnitudes and the phase angle differences within their
bounds.  Constraints \eqref{c_2d} and \eqref{c_2e} enforce the
generator active and reactive output limits.  Constraints \eqref{c_2f}
enforce the line flow limits.  Constraints \eqref{c_2g} and
\eqref{c_2h} capture \emph{Ohm's Law}. Finally, Constraint
\eqref{c_2i} and \eqref{c_2j} capture \emph{Kirchhoff's Current Law}
enforcing flow conservation at each bus. 

\begin{figure}[!t]
\centering
\parbox{1\linewidth}{\small
    \begin{mdframed}    
    \vspace{-10pt}
    \begin{flalign}
        \minimize: &\hspace{10pt}
        \sum_{i \in {\cal N}}  \text{cost}(p^g_i, \zeta_i) && \label{c_2a} \tag{2a}\\
        \text{s.t.} &\hspace{10pt}
        \bm{v}^{\min}_i \leq v_i \leq \bm{v}^{\max}_i         
                \;\; \forall i \in {\cal N}         \label{c_2b} \tag{2b}\\
        &\hspace{10pt}
        -\bm{\theta}^\Delta_{ij} \leq \theta_i - \theta_j  \leq \bm{\theta}^\Delta_{ij}   
            \;\; \forall (ij) \in {\cal E}       \label{c_2c}\!\!\!\!\! \tag{$2\bar{c}$}\\
        &\hspace{10pt}
        \bm{p}^{g\min}_i \leq p^g_i \leq \bm{p}^{g\max}_i     
            \;\; \forall i \in {\cal N}         \label{c_2d} \tag{$2\bar{d}$}\\
        &\hspace{10pt}
        \bm{q}^{g\min}_i \leq q^g_i \leq \bm{q}^{g\max}_i     
            \;\;\forall i \in {\cal N}         \label{c_2e} \tag{$2\bar{e}$}\\
        &\hspace{10pt}
        (p_{ij}^f)^2 + (q_{ij}^f)^2 \leq \bm{S}^{f\max}_{ij}           
            \;\;\forall (ij) \in {\cal E}  \label{c_2f}  \tag{$2\bar{f}$}\\
        &\hspace{10pt}
        %%%%%%%%%%%%%%%%%%%
        p_{ij}^f = \bm{g}_{ij} v_i^2 -  v_i v_j (\bm{b}_{ij} \sin(\theta_i - \theta_j) + \nonumber \\ &\hspace{33pt} \bm{g}_{ij} \cos(\theta_i - \theta_j))   
            \;\;\forall (ij)\in {\cal E} \label{c_2g} \tag{$2\bar{g}$}\\
        %%%%%%%%%%%%%%%%%%%
        &\hspace{10pt}
        q_{ij}^f = - \bm{b}_{ij} v_i^2 -  v_i v_j (\bm{g}_{ij} \sin(\theta_i - \theta_j) - \nonumber \\ &\hspace{33pt} \bm{b}_{ij} \cos(\theta_i - \theta_j))   
            \;\;\forall (ij) \in {\cal E}  \label{c_2h} \tag{$2\bar{h}$}\\
        &\hspace{10pt}
            p^g_i - \bm{p}^d_i = \textstyle \sum_{(ij)\in {\cal E}} p_{ij}^f 
            \;\;\forall i\in {\cal N}      \label{c_2i} \tag{$2\bar{i}$}\\
        &\hspace{10pt}
            q^g_i - \bm{q}^d_i = \textstyle  \sum_{(ij)\in {\cal E}} q_{ij}^f    
            \;\;\forall i\in {\cal N}      \label{c_2j} \tag{2j}\\
        %%%%%%
    \text{Output}:&\hspace{10pt} 
    (p^g, v) \text{ -- The system operational parameters}
    \!\!\!\!\!\!\!\!\!\!\!\!\!\notag
    \end{flalign}
    %\vspace{-16pt}
    \end{mdframed}
    }
    %\vspace{-6pt}
    \caption{AC Optimal Power Flow (AC-OPF).}
    \label{model:ac_opf}
    \vspace{15pt}
\end{figure}

%%%%%%%%%%%%%%%%%%%%%%%%%%%%%%%%%
\begin{figure}%[5]{r}{175pt}
\label{eq:newton_method}
%\vspace{-22pt}
\begin{mdframed}
%\vspace{-10pt}
    %{\small
    %\begin{align}
    %    \!\!\!\!\!\minimize:& \;\;
    %    \| p^g - \hat{\bm{p}}^g \|^2 + \| v - \hat{\bm{v}} \|^2 \label{load_flow_obj} 
    %    \tag{3}\\
    %    \!\!\!\!\!\text{s.t.:} & \;\; \text{Eqns.}~\ref{c_2b}-\ref{c_2j} \notag\\
    %    \text{Output}:&\;\;(p^g, v)\notag
    %\end{align}
    %}
    \begin{equation}
    \label{eq:newton_method_delta}
    \Delta \mathbf{x}_n = - \mathbf{J}^{-1}(\mathbf{x}_n) \mathbf{f}(\mathbf{x}_n)
    \end{equation}
    \begin{equation}
    \label{eq:newton_method_step}
    \mathbf{x}_{n+1} = \mathbf{x}_n + \Delta \mathbf{x}_n
    \end{equation}

%\vspace{-12pt}
\end{mdframed}
\caption{Newton's method.}
\vspace{10pt}

\label{model:load_flow}
\end{figure}
\setcounter{equation}{3} 
%%%%%%%%%%%%%%%%%%%%%%%%%%%

\paragraph{Feasibility restoration (AC-Optimal Power Flow)} Being an approximation, a LtO solution $(\hat{\bm{p}}^g, \hat{v})$ may not satisfy the original constraints. Feasibility can be restored by applying Netwon's method, which is
reported in Figure \ref{model:load_flow}. It is an iterative method that produces better approximation to the root $\mathbf{x} \in \mathbb{R}^p$, of a function $f(x) \in \mathbb{R}^m$ by iteratively solving a non-linear system of equations. If solving for $\mathbf{x}_{n+1}$, given $\mathbf{x}_{n}$, the method requires to compute the inverse of the Jacobian $J(\mathbf{x}_{n}) \in \mathbb{R}^{m\times p}$. From Eq. \ref{eq:newton_method_delta} and \ref{eq:newton_method_step}, it can be noticed that $J(\mathbf{x}_{n}) \Delta \mathbf{x}_n = - f(\mathbf{x}_{n})$, and so is possible to avoid computing the inverse of the Jacobian $J$ of $f$, and solving a linear system of equation for the unknown $\Delta \mathbf{x}_n$. In the context of restoring feasibility of the LtO solution to the AC-Optimal Power Flow problem, $f$ represents the set of inequality and equality constraint functions, from \eqref{c_2b} to \eqref{c_2h}, while $x=[v,\theta, p^g, q^g]^T$. Since the method requires each $f_i(x), i=1,\dots m$ to be an equality function, to construct a system of only equations, a $\textbf{ReLU}(f(x))=\max(0,f(x))$ is applied to each inequality function. For the AC-OPF experiment, the number of constraint function $m=602$ while the number of variables $p=472$; being $m>p$, the inverse $J^{-1}$ of the Jacobian $J$ is the generalized inverse $J^+=(J^T J)^{-1}J^T$, and $\Delta \mathbf{x}_n$ is the solution in the least square sense. The convergence of the method requires the starting point $x_0$ to be such that the $2$-norm $\lVert f(x_0) \rVert_2 \ll 1$. In the experiments, we verified that such assumption holds as evidenced by the minimal Constraint Violation achieved by each LTO method adopted (see Table \ref{table:merged_regret_violation_table}). We consider the method to have converged when the absolute value of each constraint function $|f_i(x_n)| < 1\mathrm{e}{-6}$. 

\section{Experimental Details}
\label{app:experimental}

\subsection{Portfolio Optimization Dataset}

 The stock return dataset is prepared exactly as prescribed in \cite{sambharya2023end}. The return parameters and asset prices are $\zeta = \alpha(\hat{\zeta}_t + \epsilon_t) $ where $\hat{\zeta}$ is the realized return at time $t$, $\epsilon_t$ is a normal random variable, $\epsilon_t \sim \mathcal{N}(0,\,\sigma_\epsilon I)$, and $\alpha=0.24$ is selected to minimize $\EE \| \hat{\zeta}_t-\zeta \|_2^2$. For each problem instance, the asset prices $\zeta$ are sampled by circularly iterating over the five year interval. In the experiments, see Prob. \ref{eq:opt_portfolio}, $\lambda=2.0$.

The covariance matrix $\bm{\Sigma}$ is constructed from historical price data and set as $\bm{\Sigma}= F\bm{\Sigma}_F F^T + D$, where $F \in \mathbb{R}^{n,l}$ is the factor-loading matrix, $\bm{\Sigma} \in \mathbb{S}_{+}^l$ estimates the factor returns and $D \in \mathbb{S}_{+}^l$, also called the idiosyncratic risk, is a diagonal matrix which takes into account for additional variance for each asset.

\subsection{Nonconvex Optimization Dataset}
The nonconvex optimization dataset has $2400$ samples, divided into training, validation and test set, each consisting of $2000$, $200$ and $200$ samples, respectively.
With reference to \ref{eq:bilinear} and \ref{eq:bilinear_obj}, the matrix $Q=\mu I$, where $\mu \in \mathbb{R}^n \sim \mathcal{U}(0, 1)$. The parameter $\zeta \sim \mathcal{U}(0, 5)$ and the matrix $A$ and $G$ are both drawn from the normal distribution $\mathcal{N}(0, 1)$. The right-hand side of the equality constraint $b \sim \mathcal{U}(-1, 1)$, while the right-hand side of the inequality constraint $h=\sum_{i=1}^n |M_{ij}|$, where $M=GA^+$ and $A^+=(A^T A)^{-1}A^T$.

\subsection{Nonconvex AC-OPF Dataset}
The nonconvex optimization dataset has $10000$ samples, divided into training, validation and test set, each consisting of $8334$, $833$ and $833$ samples, respectively.
The Nonconvex AC-OPF Dataset is constructed by applying random perturbations of the cost values found in NESTA benchmark case 118. More specifically, a perturbation $\mu \in \mathcal{U}(0, 100)$ is applied to each generator cost value $\zeta_i$.

\subsection{Nonconvex EPO Baselines}
The nonconvex QP variant (\ref{eq:bilinear}) of Section \ref{sec:Experiments} admits derivatives for EPO training by differentiation of the fixed-point conditions of a locally convergent solution method. Projected Gradient Descent is known to be locally convergent in nonconvex optimization \citep{attouch2013convergence}, and it is found empirically to converge locally on the problem (\ref{eq:bilinear}). 

On a problem of form
\begin{subequations}
    \label{eq:pgd_prob}
    \begin{align*}
        \bm{x}^{\star}(\bm{\zeta}) = \argmin_{\bm{x} } &\;\;
        f(\bm{x}, \bm{\zeta} ))\\
        \text{s.t.} \;\;\;\; \bm{x} \in \cS
    \end{align*}
\end{subequations}
one step of the method follows 
\begin{equation}
    \label{eq:PGD}
        \bm{x}_{k+1} =  \textit{proj}_{\cS}( \bm{x}_{k} - \alpha \nabla f(\bm{x}_{k}, \bm{\zeta} ))
\end{equation}
leading to the fixed-point conditions
\begin{equation}
    \label{eq:PGD_fixedpt}
        \bm{x}^{\star} =  \textit{proj}_{\cS}( \bm{x}^{\star} - \alpha \nabla f(\bm{x}^{\star}, \bm{\zeta} ))
\end{equation}
whose implicit differentiation results in a linear system which can be solved for $\frac{\partial \bm{x}^{\star}}{\partial \bm{\zeta}}$:
\begin{subequations}
    \label{eq:PGD_grad}
    \begin{align}
        \frac{\partial \bm{x}^{\star}}{\partial \bm{\zeta}} = \frac{\partial}{\partial \bm{x}^{\star}} \textit{proj}_{\cS}( \bm{x}^{\star} - \alpha \nabla f(\bm{x}^{\star}, \bm{\zeta} )) \cdot \frac{\partial \bm{x}^{\star}}{\partial \bm{\zeta}} \\ +   \frac{\partial}{\partial \bm{\zeta}} \textit{proj}_{\cS}( \bm{x}^{\star} - \alpha \nabla f(\bm{x}^{\star}, \bm{\zeta} ))
    \end{align}
\end{subequations}
Differentiation of the inner projection step is performed by \texttt{cvxpy} \cite{agrawal2019differentiable}, while the system (\ref{eq:PGD_grad}) is constructed and solved by \texttt{fold-opt} \cite{kotary2023folded}.

\subsection{Hyperparameters}

For all the experiments, the size of the mini-batch $\mathcal{B}$ of the training set is equal to $200$. The optimizer used for the training of the optimization proxy's is Adam, and the learning rate is chosen as the best among  $\{5e-2, 1e-2, 5e-3, 1e-3, 5e-4, 1e-4\}$.  For each task, an early stopping criteria based on the evaluation of the test-set percentage regret after restoring feasibility, is adopted to all the LtO(F) the proxies, the predictive EPO (w/o) proxy model, and pre-trained predictive model; an early stopping criteria based on the evaluation of the mean squared error is adopted to all the Two-Stage predictive model.\\  
For each optimization problem, the LtO proxies are $2$-layers ReLU neural networks with dropout equal to $0.1$ and batch normalization. All the LtOF proxies are $(k+1)$-layers ReLU neural networks with dropout equal to $0.1$ and batch normalization, where $k$ denotes the complexity of the feature mapping.
For the LtOF, Two-Stage, EPO (w/o) Proxy algorithm, the feature size of the Convex Quadratic Optimization and Non Convex AC Optimal Power Flow $\lvert z \rvert = 30$, while for the Non Convex Quadratic Optimization $\lvert z \rvert = 50$.
The hidden layer size of the feature generator model is equal to $50$, and the hidden layer size of the LtO(F) proxies, and the 2Stage, EPO and EPO w/ proxy's predictive model is equal to $500$. \\ A grid search method is adopted to tune the  hyperparameters of each LtO(F) models. For each experiments, and for each LtO(F) methods, below is reported the list of the candidate hyperparameters for each $k$, with the chosen ones marked in bold. We refer to \cite{fioretto2020lagrangian}, \cite{park2023self} and \cite{donti2021dc3} for a comprehensive description of the parameters of the LtO methods adopted in the proposed framework. 
In our result, two-stage methods report the \emph{lowest regret} found in each experiment and each $k$ across all hyperparameters adopted.

\subsubsection{Convex Quadratic Optimization and Non Convex Quadratic Optimization}

\subsection*{LD}
\begin{tabular}{ll}
\toprule
Parameter & Values \\
\midrule
$\lambda(0)$ & $\bm{0.1}$, 0.5, 1.0, 5.0, 10.0, 50.0 \\
$\mu(0)$ & 0.1, $\bm{0.5}$, 1.0, 5.0, 10.0, 50.0 \\
Training epochs & 50, 100, $\bm{200}$, 300, 500 \\
LD step size & 1.0, 0.1, 0.01, $\bm{0.001}$, 0.0001 \\
\bottomrule
\end{tabular}

\subsection*{PDL}
\begin{tabular}{ll}
\toprule
Parameter & Values \\
\midrule
$\tau$ & 0.5, 0.6, 0.7, $\bm{0.8}$, 0.9 \\
$\rho$ & 0.1, $\bm{0.5}$, 1, 10 \\
$\rho_{\text{max}}$ & 1000, $\bm{5000}$, 10000 \\
$\alpha$ & 1, 1.5, 2.5, $\bm{5}$, 10 \\
\bottomrule
\end{tabular}

\subsection*{DC3}
\begin{tabular}{ll}
\toprule
Parameter & Values \\
\midrule
$\lambda+\mu$ & 0.1, 1.0, $\bm{10.0}$, 50.0, 100.0 \\
$\frac{\lambda}{\lambda+\mu}$ & 0.1, 0.5, $\bm{0.75}$, 1 \\
$t_{\text{test}}$ & 1, 2, $\bm{5}$, 10, 100 \\
$t_{\text{train}}$ & 1, 2, $\bm{5}$, 50, 100 \\
\bottomrule
\end{tabular}

\subsubsection*{Non Convex AC-Optimal Power Flow }
\subsection*{LD}
\begin{tabular}{ll}
\toprule
Parameter & Values \\
\midrule
$\lambda(0)$ & $\bm{0.1}$, 0.5, 1.0, 5.0, 10.0, 50.0 \\
$\mu(0)$ & $\bm{0.1}$, 0.5, 1.0, 5.0, 10.0, 50.0 \\
Training epochs & 50, 100, 200, $\bm{300}$, 500 \\
LD step size & $\bm{1.0}$, 0.1, 0.01, 0.001, 0.0001 \\
\bottomrule
\end{tabular}

\subsection*{PDL}
\begin{tabular}{ll}
\toprule
Parameter & Values \\
\midrule
$\tau$ & 0.5, 0.6, 0.7, $\bm{0.8}$, 0.9 \\
$\rho$ & 0.1, 0.5, $\bm{1}$, 10 \\
$\rho_{\text{max}}$ & 1000, 5000, $\bm{10000}$ \\
$\alpha$ & 1, 1.5, 2.5, 5, $\bm{10}$ \\
\bottomrule
\end{tabular}

\iffalse
\subsection{Training time}

\begin{table}[h]
\centering
\caption{Total Training Time for Methods}
\label{table:training_times}
\resizebox{0.5\textwidth}{!}{
\begin{tabular}{l|c|c|c|c|c}
\hline
\textbf{Method} & \textbf{PDL} & \textbf{LD} & \textbf{DC3} & \textbf{2-Stg} & \textbf{EPO} \\ \hline
ACOPF       & 1211.1 s & 1048.7 s & - & 46.5 s      & -        \\ 
PORTFOLIO   & 1174.2 s & 1023.8 s & 1281.1 s & 125.9 s & 1563.6 s \\
NONCONVEXQP & 1364.3 s & 1316.6 s & 1449.1 s & 86.2 s & 2582.2 s \\ \hline
\end{tabular}}
\end{table}
\fi

\end{document}